# Generalised framework for multi-criteria method selection☆

Jarosław Wątróbski[a,*], Jarosław Jankowski[b], Paweł Ziemba[a], Artur Karczmarczyk[b], Magdalena Zioło[a]

[a] Faculty of Economics and Management, University of Szczecin, Mickiewicza 64, 71-101 Szczecin, Poland
[b] Faculty of Computer Science and Information Systems, West Pomeranian University of Technology, Zolnierska 49, 71-210 Szczecin, Poland

ARTICLE INFO

ABSTRACT

Multi-Criteria Decision Analysis (MCDA) methods are widely used in various fields and disciplines. While most of the research has been focused on the development and improvement of new MCDA methods, relatively limited attention has been paid to their appropriate selection for the given decision problem. Their improper application decreases the quality of recommendations, as different MCDA methods deliver inconsistent results. The current paper presents a methodological and practical framework for selecting suitable MCDA methods for a particular decision situation. A set of 56 available MCDA methods was analysed and, based on that, a hierarchical set of methods' characteristics and the rule base were obtained. This analysis, rules and modelling of the uncertainty in the decision problem description allowed to build a framework supporting the selection of a MCDA method for a given decision-making situation. The practical studies indicate consistency between the methods recommended with the proposed approach and those used by the experts in reference cases. The results of the research also showed that the proposed approach can be used as a general framework for selecting an appropriate MCDA method for a given area of decision support, even in cases of data gaps in the decision-making problem description. The proposed framework was implemented within a web platform available for public use at www.mcda.it.

## 1. Introduction

The increasing complexity of economic and social systems results in an increase in the complexity of the related decision problems [1]. They concern, among others, political decisions [2], organization management [3,4], financial management [5,6] and marketing [7]. Many decision problems are characterized by large dimensionality [8], the occurrence of sources of uncertainty and risk factors [9]. It is important to reconcile the contradictory goals, make decisions with many criteria and strive for compromise solutions [10]. Policy makers face the complexity of decision situations and they require methods and systems that support the decision-making [11]. In response to these needs, many solutions dedicated to selected areas, as well as general-purpose methods have been developed [12]. In this context, multi-criteria decision analysis (MCDA) methods are widely used. Apart from the formal foundations, these methods are characterized by the possibilities of handling a multitude of conflicting goals, as well as different stakeholders within decision making process [13]. In the recent years, a dynamic development of MCDA methods has been observed [14]. However, they significantly differ in many dimensions such as complexity, the way in which preferences and evaluation criteria are represented, the type of data aggregation, the possibility of including uncertain data, and the availability of implementations in decision support systems or criteria compensation [15–17]. The extensive number of possible MCDA methods results in a problem with their proper selection and application in specific decision situations.

While various MCDA methods can be used for improving the quality of decisions, they often produce conflicting results when compared [18–22]. It is worth noticing that a decision-maker (DM) may reach different decisions even when applying the same weights of criteria and the criterial evaluations of variants. This fact has been confirmed in a number of publications, in which rankings of decision variants were examined with the use of different MCDA methods [1,19,23–26]. Such analytical methods also often fail to provide guidelines [16]. The question is, which

☆ This manuscript was processed by Associate Editor Prof. B. Lev.

* Corresponding author.
*E-mail addresses:* jaroslaw.watrobski@wneiz.pl (J. Wątróbski), jjankowski@wi.zut.edu.pl (J. Jankowski), pziemba@ajp.edu.pl (P. Ziemba), akarczmarczyk@wi.zut.edu.pl (A. Karczmarczyk), magdalena.ziolo@usz.edu.pl (M. Zioło).

decision support method should be used and which characteristics of the decision problems are affecting the selection of a method?

A significant research issue, which is not entirely solved yet, is to determine a method suitable for a given problem, since only a method which is correctly chosen allows to obtain a solution that is most satisfying for the DM [20] in the context of a given decision-making situation. This problem emerges when the decision maker is unable to obtain a detailed description of the decision-making situation [18,27]. The complexity, uniqueness, or the fact that the decision-making situations can occur simultaneously in a short period makes analysis of them challenging [20,28]. In such conditions, the DM faces a dilemma of either making the decision based on incomplete information, or not making it on time [29]. In consequence, it becomes necessary to use formal procedures and guidelines for selecting MCDA methods also in cases of partial lack of knowledge about the decision-making situation.

The literature review provides a vast range of works dealing with the MCDA method selection problem for a given decision-making problem. However, the range of these solutions is often limited to the few of the best-known MCDA methods [30,31] or to a single, arbitrarily selected, field of application [22,32]. The studied approaches often also require that a decision-maker knows in advance certain formal aspects of the problems. In reality, a decision-maker may find it difficult to define a priori all relevant details of a given decision situation. Unfortunately, there is a lack of approaches addressing this uncertainty.

As a result, the motivation of the current research was:

- to build a formal guideline for MCDA method selection, which is independent of the problem domain,
- to use an extensive set of available MCDA methods and their characteristics,
- to obtain high accuracy of recommendation of particular MCDA methods for a given decision-making situation,
- to address the lack of knowledge issue in the descriptions of the decision-making situations.

In this paper, a new approach for selecting an MCDA method is proposed. As the authors aimed to develop an approach independent of the area of usage, the proposed framework is based on determining a set of characteristics of the available MCDA methods. Furthermore, the authors endeavoured to address the knowledge gaps in decision-making situation description and, additionally, to analyse their influence on the process of the MCDA method selection. The authors' technical contribution is also provided in a form of a useful website-based tool for supporting the process of MCDA method selection.

According to the authors' best knowledge, this research is the first successful attempt to handle uncertainty in the decision-making situation description during MCDA methods selection process. The entire solution space was examined. Surprisingly, the results clearly show that even partial uncertainty in a selected aspect of the decision-making situation description does not significantly affect the contents of the recommended set of methods.

The practical confirmation of the proposed framework was based on scientific literature as a reliable source of expert knowledge and the fact that usually the decision makers, who are experts, have knowledge of which method should provide a sufficient solution to the problem [33]. Practical examples are positioned in the field of sustainable logistics and transport as the field with wide usage of MCDA methods [30,34]. Research confirmed that the recommendations for MCDA methods' usage delivered by the proposed framework are consistent with the methods used by the experts for solving specific problems.

The paper is organized as follows: Section 2 provides the MCDA methodology foundations followed by the definition of the research gap. In Section 3, a framework of multi-criteria method selection is provided. A discussion of the range in which uncertainty of the decision-making problem description affects the framework is also presented. An outline of an expert system supporting MCDA methods selection is also provided. In Section 4, an exemplary confirmation of the proposed framework in the area of sustainable transport and logistics is presented. The article concludes with a discussion of the achieved results and areas of further research.

## 2. Literature review

### 2.1. MCDA foundations

The MCDA methods' task is to support a decision-maker in choosing the most preferable variant from many possible options, taking into account a multitude of criteria characterizing acceptability of individual decision variants. The criteria can also grade the quality of the variants when all options are permissible and the problem is to choose the best one subjectively. In this case, subjectivity refers to the importance of individual criteria, as for each decision-maker some factors are typically more significant than others. Furthermore, the uncertainty and inaccuracy of data describing alternatives influence the subjectivity of evaluation [35].

Multi-criteria problems can be divided into continuous ones, such as multiple-criteria linear programming, and discrete ones, such as those solved by methods based on utility or value function and outranking methods [36]. The utility/value theory-based approach determines two types of relationships between variants: indifference ($a_i \ I \ a_j$) and preference ($a_i \ P \ a_j$) of one variant over another. Methods in this group leave out non-comparability of the decision variants and assume transitivity and completeness of preference [29]. Methods based on outranking relations often expand a set of basic preferential situations with the result that contains indifference of decision variants ($a_i \ I \ a_j$), weak preference - one variant over another ($a_i \ Q \ a_j$), the strict preference - one variant over another ($a_i \ P \ a_j$), and incomparability between data variations ($a_i \ R \ a_j$) [29]. The preferential situations can be combined in an “outranking” relation, which contains the situations of indifference as well as strict and weak preference ($a_i \ S \ a_j$) [37].

The preference scenarios in the outranking methods are related to the thresholds used in them (outranking methods). Indifference ($q$), preference ($p$) and veto ($v$) are the three kinds of thresholds [27]. The thresholds allow the recognition of the uncertainty of the evaluations by the preferences' gradation. Furthermore, in many outranking methods (e.g. ELECTRE III), the weak preference has the form of a linear function whose values, from the interval *[0, 1]*, increase when approaching the threshold $p$, and, as a result, the preferences are subject to a characteristic fuzzification. Moreover, the preference thresholds' usage determines the form of the preference criterion used in the MCDA method. When no thresholds are used, the MCDA method uses a so-called true-criterion. However, application of the indifference threshold only determines the use of a semi-criterion by the method, and application of the indifference and preference thresholds means that the method uses a pseudo-criterion [38].

Two basic operational approaches may be distinguished to aggregate performance of variants: (1) aggregation to a single criterion (American school), (2) aggregation by using the outranking relationship (European school) [37]. Moreover, mixed (indirect) approaches, which combine elements of American and European decision-making schools, are applied. The approach can be exemplified by a group of PCCA (Pairwise Criterion Comparison Approach) methods [39].

MCDA methods are also different depending on the nature and characteristics of the used data [28]. The nature of data is closely connected to the measurement scale. Data can be quantitative or qualitative and can be expressed in the cardinal (quantitative) or ordinal (qualitative) scale [40]. What is more, the cardinal scale can be of interval or ratio (relative) type [35]. In case of a relative scale, the data is presented in relations to other data. For example, the weight of criterion $g_1$ can be expressed in relations to criterion $g_2$ ($g_1$ is 3 times more important than $g_2$) [57]. The characteristics of the data used refer to whether the data is certain or uncertain [41]. The certain data, which is also called deterministic, is expressed in a crisp form, whereas uncertain data (non-deterministic) is represented by some kind of distribution (continuous or discrete) [20,28]. New methods based on the fuzzy set theory make it possible to express uncertain data in a fuzzy form [41]. The data type refers to both the scale on which the criterion performance of the variants is presented, as well as to the weights of the criteria. A summary of individual MCDA methods and their basic properties is presented in Table 1 and Supplementary material – Section 1.

According to Roy, there are four stages in the decision-making process [37]: (I) defining an object of the decision and the set of potential decision variants A as well as the determination of the reference problematics on A; (II) analysing consequences and developing the consistent set of criteria C; (III) modelling comprehensive preferences and operationally aggregating performances; (IV) investigating and developing the recommendation, based on the results of stage III and the problem defined in Stage I. Roy argues that the stages are not serial. For instance, some elements of Stage I can require performing elements of Stage II. Similarly, the decision-making process cannot be simplified by eliminating individual stages. In Stage I, Roy [35] distinguishes four decision problematics: $\alpha$ - selection, $\beta$ - sorting, $\gamma$ - ranking, $\delta$ - description with formal representation presented in Supplementary material – Section 2.

In Stage III, the operational approach for a given decision problem should be selected. Stage IV, in particular, requires selecting the computational procedure (the MCDA method), depending on the decision issue and the decision-maker's operational approach [37]. Roy's model indicates that the selection of the MCDA method is a vital element of solving a decision problem [17]. Furthermore, to obtain a "good" solution to the problem, one needs to apply a properly selected method.

### 2.2. The problem of selection of a proper MCDA method

Even in the early study of [18], it was found that "the great diversity of MCDA procedures may be seen as a strong point, it can also be a weakness. Up to now, there has been no possibility of deciding whether one method makes more sense than another in a specific problem situation. A systematic axiomatic analysis of decision procedures and algorithms is yet to be carried out."

Roy [37] also indicated that the selection of the MCDA method is a vital element of solving a decision problem. When defining the operational approach, the author paid attention to the method selection problem within four stages in the decision-making process. Furthermore, to obtain a "good" solution to the problem, a decision-maker needs to apply an adequately selected method [17]. However, selecting a multi-criteria method only on the basis of the decision issue and operational approach seems to be too general, as the decision-maker can choose many methods to solve a given decision problem on such a basis. The issue is the multitude of MCDA methods and their diversity [33,42].

Decision-makers are often unable to fully justify their choice of the method which was applied to solve their decision situation [21]. The selection of a multi-criteria method is usually carried out arbitrarily and is motivated by the decision-maker's knowledge of a given method or availability of software supporting the method [10,43]. Similar issues are also levelled in relation to MCDA software selection. Decision-makers usually choose decision support software, which, they are familiar with [44]. On this account, it is not an MCDA method that is selected for a decision problem, but the decision problem is adjusted to a chosen multi-criteria method [20]. It is difficult to answer a question which method is most suitable to solve a given kind of a problem [18,19]. The selection of a proper MCDA method for a given decision situation is salient, since various methods can yield different results for the same problem [18–26]. The difference in results when applying various calculating procedures can be influenced by the following factors [19,45]: (a) various techniques use weights differently in their calculations; (b) algorithms differ in their approach to selecting the "best" solution; (c) many algorithms attempt to scale the objectives, which affects the weights already chosen; (d) some algorithms introduce additional parameters affecting the final recommendations.

The literature analysis shows several works dealing with the subject of multi-criteria method selection for a given decision problem. They can be categorized into those which, when selecting an MCDA method, were based on: benchmarking [1,19,25,26,46,47], multi-criteria methods (it was recognized that the issue of selecting an MCDA method is a multi-criteria problem) [48] as well as the informal [16,30,49] or formal [21,22,31,32,50,51,52] structuring of a problem or a decision situation. A summary of the up to date approaches to MCDA method selection is presented in Supplementary material – Section 3. The presented approaches are not without shortcomings. The benchmark-based approaches ignore that the solutions considered for decision-making are usually optimal in Pareto term. In fact, they do not allow them to choose the optimal MCDA method, but only compare the compliance of the solutions of each method. The multi-criteria approach places the problem of methods selection in loop as it requires the use of MCDA method [20,47]. In turn, the informal approach does not give clear and unambiguous guidance on the choice of the method of MCDA, applicable to the particular class of decision problem. The formal approaches are characterized by an accurate selection of the methods oscillating on the border of acceptability. The IDEA approach [21,22,32] achieves an accuracy of 63–73%, depending on the matched MCDA method. The range of discipline and methodical approaches used so far, often limiting them only to the analysed domain of MCDA methods usage, is also problematic.

The guidelines for the selection of the MCDA method may be redundant for some classes of decision problems. The degree of criteria compensation is essential for the problems in the field of sustainability [30], but for other classes of problems, such a guideline is unnecessary. When analysing the coverage of individual methodical approaches, it should be noted that the previously mentioned works on the MCDA method selection considered a comparatively limited set of methods. The highest number of 29 methods was examined in [32,51] considered the 24 methods, [21] - 22, [22] - 21, [25] - 18, [48] - 16, [19] - 8, [50] - 6, [30] - 5, [31] - 4, [52] - 3, [26] - 3, [1] - 3. The publications often failed to include the relatively new methods such as ANP, Vikor and fuzzy extensions of the top classical methods. As a result the motivation of the current research was to build a formal guideline and framework for MCDA method selection independent from the problem domain with the use of a complete set of available MCDA methods and their characteristics.

## 3. The proposed framework for MCDA method selection

### 3.1. Main assumptions

In this section, we propose a generalized framework for the selection of a suitable MCDA method for a particular decision situation. The conceptual framework is shown in Fig. 1. There are two separate components of the framework, the methodological and the practical elements. The first one is based on methodological aspects and the rules' database generation. Both elements are required for the second part, i.e. practical verification when the descriptors of the considered problem are gathered and the subset of recommended methods is presented. The methodological aspects include creating a set of the considered MCDA methods and an analysis of their properties. The set of characteristics for each method is obtained and presented in Table 2. The subset of the recommended MCDA methods is obtained on the basis of the rule base (Table 3) and the characteristics of the problem considered. The validity of the proposed framework is reported in the following section.

**Table 1**
Taxonomy of MCDA methods.

| Method name | Available binary relations | | | | | Linear compensation effect | | | Type of aggregation | | | Type of preferential information | | | | |
|---|---|---|---|---|---|---|---|---|---|---|---|---|---|---|---|---|
| | *I* | *P* | *Q* | *R* | *S* | No | Total | Partial | Single criterion | Outranking | Mixed | Deterministic | Cardinal | Non-deterministic | Ordinal | Fuzzy |
| AHP | 1 | 1 | 0 | 0 | 0 | 0 | 0 | 1 | 1 | 0 | 0 | 1 | 1 | 1 | 0 | 0 |
| ANP | 1 | 1 | 0 | 0 | 0 | 0 | 0 | 1 | 1 | 0 | 0 | 1 | 1 | 1 | 0 | 0 |
| ARGUS | 1 | 0 | 0 | 1 | 1 | 0 | 0 | 1 | 0 | 1 | 0 | 1 | 0 | 1 | 1 | 0 |
| COMET | 1 | 1 | 0 | 0 | 0 | 0 | 1 | 0 | 1 | 0 | 0 | 1 | 1 | 1 | 1 | 1 |
| ELECTRE I | 0 | 0 | 0 | 1 | 1 | 0 | 0 | 1 | 0 | 1 | 0 | 1 | 1 | 0 | 1 | 0 |
| ELECTRE II | 0 | 0 | 0 | 1 | 1 | 0 | 0 | 1 | 0 | 1 | 0 | 1 | 1 | 0 | 1 | 0 |
| ELECTRE III | 0 | 0 | 0 | 1 | 1 | 0 | 0 | 1 | 0 | 1 | 0 | 1 | 1 | 0 | 1 | 0 |
| ELECTRE IS | 0 | 0 | 0 | 1 | 1 | 0 | 0 | 1 | 0 | 1 | 0 | 1 | 1 | 0 | 1 | 0 |
| ELECTRE IV | 0 | 0 | 0 | 1 | 1 | 0 | 0 | 1 | 0 | 1 | 0 | 1 | 1 | 0 | 1 | 0 |
| ELECTRE TRI | 0 | 0 | 0 | 1 | 1 | 0 | 0 | 1 | 0 | 1 | 0 | 1 | 1 | 0 | 1 | 0 |
| EVAMIX | 1 | 1 | 0 | 0 | 0 | 0 | 0 | 1 | 1 | 0 | 0 | 1 | 1 | 0 | 1 | 0 |
| Fuzzy AHP | 1 | 1 | 0 | 0 | 0 | 0 | 0 | 1 | 1 | 0 | 0 | 1 | 1 | 1 | 0 | 1 |
| Fuzzy ANP | 1 | 1 | 0 | 0 | 0 | 0 | 0 | 1 | 1 | 0 | 0 | 1 | 1 | 1 | 0 | 1 |
| Fuzzy methods of extracting the minimum and maximum values of the attribute | 1 | 1 | 1 | 0 | 0 | 1 | 0 | 0 | 0 | 0 | 1 | 0 | 1 | 1 | 1 | 1 |
| Fuzzy PROMETHEE I | 1 | 1 | 0 | 1 | 0 | 0 | 0 | 1 | 0 | 1 | 0 | 1 | 1 | 1 | 1 | 1 |
| Fuzzy PROMETHEE II | 1 | 1 | 0 | 0 | 0 | 0 | 0 | 1 | 0 | 1 | 0 | 1 | 1 | 1 | 1 | 1 |
| Fuzzy SAW | 1 | 1 | 1 | 0 | 0 | 0 | 1 | 0 | 1 | 0 | 0 | 0 | 1 | 1 | 1 | 1 |
| Fuzzy TOPSIS | 1 | 1 | 0 | 0 | 0 | 0 | 1 | 0 | 1 | 0 | 0 | 1 | 1 | 1 | 0 | 1 |
| Fuzzy VIKOR | 1 | 1 | 0 | 0 | 0 | 0 | 1 | 0 | 1 | 0 | 0 | 1 | 1 | 1 | 0 | 1 |
| IDRA | 1 | 1 | 0 | 0 | 0 | 0 | 0 | 1 | 0 | 0 | 1 | 1 | 1 | 1 | 0 | 0 |
| Lexicographic method | 1 | 1 | 0 | 0 | 0 | 1 | 0 | 0 | 1 | 0 | 0 | 1 | 1 | 0 | 1 | 0 |
| MACBETH | 1 | 1 | 0 | 0 | 0 | 0 | 0 | 1 | 1 | 0 | 0 | 1 | 1 | 1 | 1 | 0 |
| MAPPAC | 1 | 1 | 1 | 1 | 0 | 0 | 0 | 1 | 0 | 0 | 1 | 1 | 1 | 0 | 0 | 0 |
| MAUT | 1 | 1 | 0 | 0 | 0 | 0 | 0 | 1 | 1 | 0 | 0 | 0 | 1 | 1 | 0 | 0 |
| MAVT | 1 | 1 | 0 | 0 | 0 | 0 | 0 | 1 | 1 | 0 | 0 | 1 | 1 | 0 | 0 | 0 |
| Maximax | 1 | 1 | 0 | 0 | 0 | 1 | 0 | 0 | 1 | 0 | 0 | 1 | 1 | 0 | 1 | 0 |
| Maximin | 1 | 1 | 0 | 0 | 0 | 1 | 0 | 0 | 1 | 0 | 0 | 1 | 1 | 0 | 1 | 0 |
| Maximin fuzzy method | 1 | 1 | 1 | 0 | 0 | 1 | 0 | 0 | 1 | 0 | 0 | 1 | 1 | 1 | 1 | 1 |
| MELCHIOR | 0 | 0 | 0 | 1 | 1 | 0 | 0 | 1 | 0 | 1 | 0 | 1 | 0 | 0 | 1 | 0 |
| Methods of extracting the minimum and maximum values of the attribute | 1 | 1 | 0 | 0 | 0 | 1 | 0 | 0 | 0 | 0 | 1 | 1 | 1 | 0 | 1 | 0 |
| NAIADE I | 0 | 0 | 0 | 1 | 1 | 0 | 0 | 1 | 0 | 1 | 0 | 1 | 1 | 1 | 1 | 1 |
| NAIADE II | 0 | 0 | 0 | 0 | 1 | 0 | 0 | 1 | 0 | 1 | 0 | 1 | 1 | 1 | 1 | 1 |
| ORESTE | 1 | 1 | 0 | 1 | 0 | 0 | 0 | 1 | 0 | 1 | 0 | 1 | 0 | 0 | 1 | 0 |
| PACMAN | 1 | 1 | 1 | 1 | 0 | 0 | 0 | 1 | 0 | 0 | 1 | 1 | 1 | 0 | 1 | 0 |
| PAMSSEM I | 0 | 0 | 0 | 1 | 1 | 0 | 0 | 1 | 0 | 1 | 0 | 1 | 1 | 1 | 1 | 1 |
| PAMSSEM II | 0 | 0 | 0 | 0 | 1 | 0 | 0 | 1 | 0 | 1 | 0 | 1 | 1 | 1 | 1 | 1 |
| PRAGMA | 1 | 1 | 0 | 1 | 0 | 0 | 0 | 1 | 0 | 0 | 1 | 1 | 1 | 0 | 0 | 0 |
| PROMETHEE I | 1 | 1 | 0 | 1 | 0 | 0 | 0 | 1 | 0 | 1 | 0 | 1 | 1 | 0 | 1 | 0 |
| PROMETHEE II | 1 | 1 | 0 | 0 | 0 | 0 | 0 | 1 | 0 | 1 | 0 | 1 | 1 | 0 | 1 | 0 |
| QUALIFLEX | 0 | 0 | 0 | 1 | 1 | 0 | 0 | 1 | 0 | 0 | 1 | 1 | 0 | 0 | 1 | 0 |
| REGIME | 0 | 0 | 0 | 1 | 1 | 0 | 0 | 1 | 0 | 1 | 0 | 1 | 0 | 0 | 1 | 0 |
| Simple Additive Weighting (SAW) | 1 | 1 | 0 | 0 | 0 | 0 | 1 | 0 | 1 | 0 | 0 | 1 | 1 | 0 | 0 | 0 |
| SMART | 1 | 1 | 0 | 0 | 0 | 0 | 0 | 1 | 1 | 0 | 0 | 1 | 1 | 0 | 0 | 0 |
| TACTIC | 1 | 1 | 0 | 1 | 0 | 0 | 0 | 1 | 0 | 1 | 0 | 1 | 1 | 1 | 0 | 0 |
| TOPSIS | 1 | 1 | 0 | 0 | 0 | 0 | 1 | 0 | 1 | 0 | 0 | 1 | 1 | 0 | 0 | 0 |
| UTA | 1 | 1 | 0 | 0 | 0 | 0 | 0 | 1 | 1 | 0 | 0 | 1 | 0 | 0 | 1 | 0 |
| VIKOR | 1 | 1 | 0 | 0 | 0 | 0 | 1 | 0 | 1 | 0 | 0 | 1 | 1 | 0 | 0 | 0 |
| DEMATEL | 1 | 1 | 0 | 0 | 0 | 0 | 1 | 0 | 1 | 0 | 0 | 1 | 0 | 1 | 1 | 0 |
| REMBRANDT | 1 | 1 | 0 | 0 | 0 | 0 | 0 | 1 | 1 | 0 | 0 | 1 | 1 | 1 | 0 | 0 |

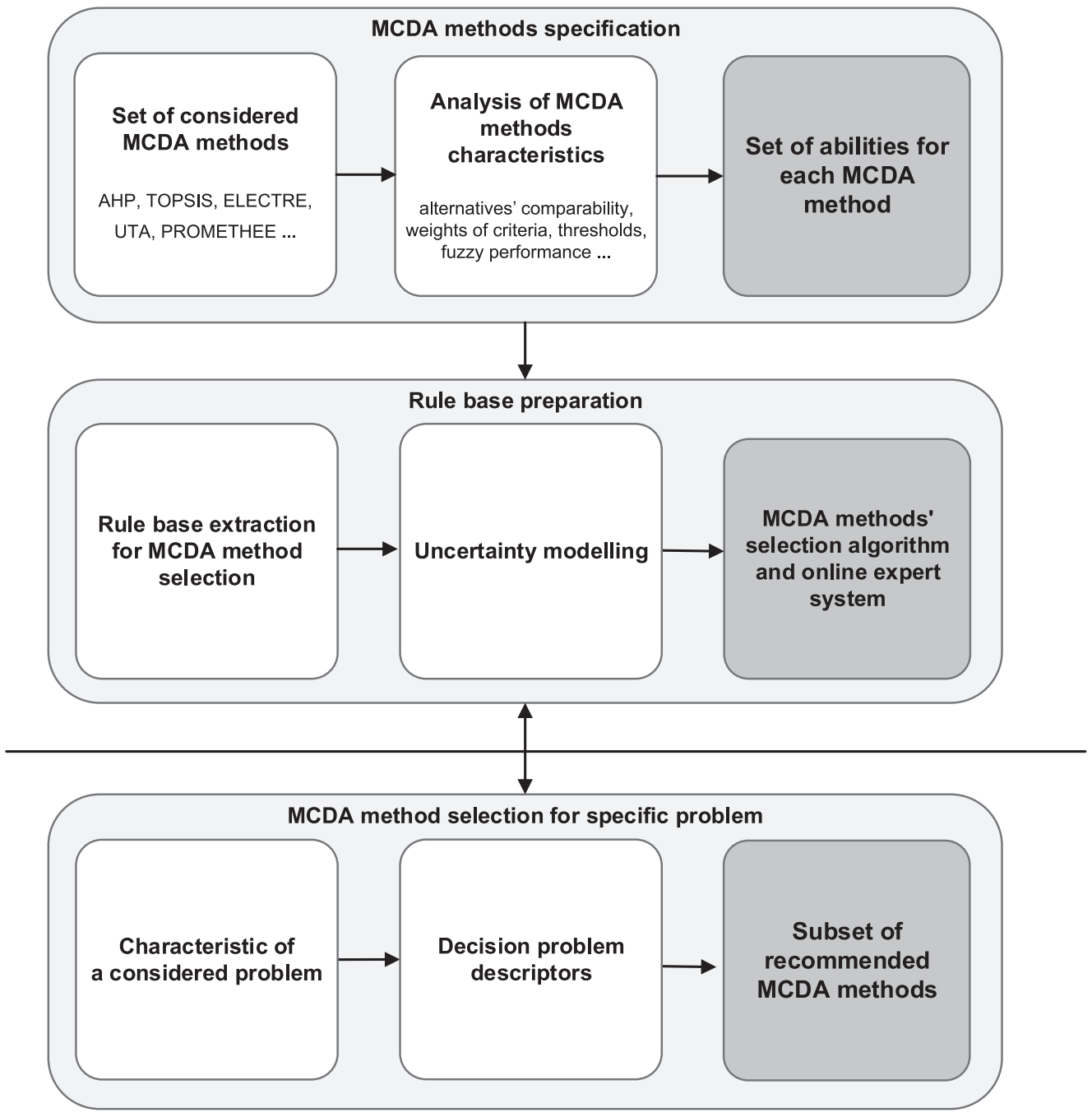

**Fig. 1.** Research procedure.

Let *DP* be a specific multi-criteria decision problem. The classic approach to the decision problem allows presenting it in the form of a three-element set ($A$, $G$, $E$), where $A$ defines a set of decision variants; $G$ is a set of criteria, and $E$ represents the efficiency of criterial performance, wherein $E = G(A)$ [20,35,51]. If sets $A$ and $G$ are presented as vectors, then set $E$ is given as a matrix $E = A^T \cdot G$. Additionally, an aggregated performance of variants can be presented in short as $E(A)$. The weights of criteria W can be defined absolutely, e.g. $g_1$, or in respect to other criteria, e.g. $g_1/g_2$. In a similar way, the criteria performances of variants can be expressed, e.g. $g_1(a_1)$ or $g_1(a_1)/g_1(a_2)$. New MCDA methods often use the fuzzy set theory [14], which allows using uncertain data as trapezoidal fuzzy numbers, i.e. $\tilde{N} = (n_l, n_u, \alpha_F, \beta_F)$, or triangular fuzzy numbers, i.e. $\tilde{N} = (n, \alpha_F, \beta_F)$ [53-55].

The basis of the proposed framework is a set of 56 MCDA methods (or their combinations) and their characteristics containing nine descriptive properties of MCDA methods organized in a hierarchical form. It is worth noting that the decision-maker does not always have full knowledge of the given decision problem. Therefore, in certain situations, the DM is not able to fully define the descriptors ($c$) of the decision problem and, therefore, also his needs regarding the characteristics ($m$) of the MCDA method. For this reason, we propose a hierarchical structure of descriptors and characteristics, adapted to various levels of the definition of the DM's needs. Each level of hierarchy is deepening the accuracy of the description from the preceding level.

### 3.2. Proposed decision problem descriptors and MCDA methods' properties

An $i$-element set $M$ of MCDA methods and a vector $m$ of their properties with a dimension $\dim(m)$ are given. Therefore, there exists a matrix describing the properties of individual methods, in a form of $TAB$ with dimensions $i \times \dim(m)$. There is also a decision-making problem (DP) described by decision problem descriptors in the form of a vector $c$, with $\dim(c) \leq \dim(m)$. For a given decision-making problem, from the set (vector) $c$, a subset of descriptors $\tilde{c}$ constituting a description of the decision-making situation is determined: $f(DP, c) = \tilde{c}$, where $\tilde{c} = [\tilde{c}_1, \ldots, \tilde{c}_j]$. Having a subset $\tilde{c}$ in the form of a vector and having a matrix $TAB$ of the MCDA methods properties, a subset $\tilde{M}_{TAB}$ of methods is constructed according to the formula:

$$\tilde{M}_{TAB} = \left\{ M_k : \ \forall_{l \in [1,j]} \ \exists_{x \in [1, \dim(m)]} \ \tilde{c}_l = TAB[k, x] \ for \ 1 \leq k \leq i \right\}$$

where $i$ denotes the number of MCDA methods in the set $M$, $j$ denotes the number of descriptors in the subset $\tilde{c}$ (the length of the vector $\tilde{c}$).

The description of the decision-making problem, expressed with the $c$ descriptors, is in accordance with the subset of the vector $m$ of the properties of the MCDA methods belonging to the set $M$. Although formally the decision-making situation descriptors and the MCDA method properties are different sets, the problem descriptors are accurately reflected by the properties of particular methods.

#### 3.2.1. Decision problem descriptors

In the proposed framework, we show that, each decision-making problem can be described by the DM using the maximum of nine descriptors belonging to the set $\tilde{c} \subseteq c$.

At the first level of the hierarchy, the DM only defines the general descriptors of the decision problem:

- c1 – whether different weights of the individual criteria will be taken into account in the decision problem; possible values are: 0 – no, 1 – yes;
- c2 – on what scale the criterial performance of the variants will be compared; possible values are: 1 – qualitative, 2 – quantitative, 3 – relative;
- c3 – whether the decision problem is characterized by uncertainty; possible values are: 0 – no, 1 – yes;
- c4 – what the decision problematic is; possible values are: 1 – selection, 2 – classification, 3 – ranking+selection,[1] 4 – classification+selection.

Of course, the knowledge about the decision problem can be clarified by the DM. While we can assume that $c_2$ is fully defined, the rest of the descriptors of the decision problem on the second level of the proposed hierarchy are presented as follows:

- c1.1 – if weights are used, what their type will be; possible values are: 1 – qualitative, 2 – quantitative, 3 – relative;
- c3.1 – if the problem is characterized by uncertainty, which uncertainty aspect it concerns; possible values are: 1 – input data uncertainty, 2 – DM's preference uncertainty, 3 – both;
- c4.1 – if the problematic of ranking is considered, what kind of variants' ranking is expected; possible values are: 1 – partial ranking, 2 – complete ranking.

The third level of the descriptors' hierarchy refers only $c_{3.1}$ and addresses data or preference uncertainty in the decision problem:

[1] The MCDA methods which deal with the ranking problematic are also efficient when considering the issue of choice

**Table 2**
The set of properties of the considered MCDA methods.

| $M_i$ | MCDA method | Abbr. | $m_{i1}$ | $m_{i1.1}$ | $m_{i2}$ | $m_{i3}$ | $m_{i3.1}$ | $m_{i3.1.1}$ | $m_{i3.1.2}$ | $m_{i4}$ | $m_{i4.1}$ | Reference |
|---|---|---|---|---|---|---|---|---|---|---|---|---|
| $\mathbf{M_1}$ | AHP | $A_H$ | 1 | 3 | 3 | 0 | 0 | 0 | 0 | 3 | 2 | [56] |
| $\mathbf{M_2}$ | AHP + TOPSIS | $A_H + T_P$ | 1 | 3 | 2 | 0 | 0 | 0 | 0 | 3 | 2 | [56] |
| $\mathbf{M_3}$ | ANP | $A_N$ | 1 | 3 | 3 | 0 | 0 | 0 | 0 | 3 | 2 | [57] |
| $\mathbf{M_4}$ | ARGUS | $A_G$ | 1 | 1 | 1 | 0 | 0 | 0 | 0 | 1 | 0 | [58] |
| $\mathbf{M_5}$ | COMET | $C_T$ | 0 | 0 | 2 | 1 | 1 | 2 | 0 | 3 | 2 | [59] |
| $\mathbf{M_6}$ | ELECTRE I | $E_1$ | 1 | 2 | 1 | 0 | 0 | 0 | 0 | 1 | 0 | [27] |
| $\mathbf{M_7}$ | ELECTRE II | $E_2$ | 1 | 2 | 1 | 0 | 0 | 0 | 0 | 3 | 1 | [27] |
| $\mathbf{M_8}$ | ELECTRE III | $E_3$ | 1 | 2 | 2 | 1 | 2 | 0 | 3 | 3 | 1 | [60] |
| $\mathbf{M_9}$ | ELECTRE IS | $E_S$ | 1 | 2 | 2 | 1 | 2 | 0 | 3 | 1 | 0 | [27] |
| $\mathbf{M_{10}}$ | ELECTRE IV | $E_4$ | 0 | 0 | 1 | 1 | 2 | 0 | 3 | 3 | 1 | [27] |
| $\mathbf{M_{11}}$ | ELECTRE TRI | $E_T$ | 1 | 2 | 2 | 1 | 2 | 0 | 3 | 2 | 0 | [27] |
| $\mathbf{M_{12}}$ | EVAMIX | $E_V$ | 1 | 2 | 2 | 0 | 0 | 0 | 0 | 3 | 2 | [61] |
| $\mathbf{M_{13}}$ | Fuzzy AHP | $A_F$ | 1 | 3 | 3 | 1 | 1 | 3 | 0 | 3 | 2 | [62] |
| $\mathbf{M_{14}}$ | Fuzzy AHP + fuzzy TOPSIS | $A_F + T_F$ | 1 | 3 | 2 | 1 | 1 | 3 | 0 | 3 | 2 | [63] |
| $\mathbf{M_{15}}$ | Fuzzy ANP | $A_{NF}$ | 1 | 3 | 3 | 1 | 1 | 3 | 0 | 3 | 2 | [64] |
| $\mathbf{M_{16}}$ | Fuzzy ANP + fuzzy TOPSIS | $A_{NF} + T_F$ | 1 | 3 | 2 | 1 | 1 | 3 | 0 | 3 | 2 | [56] |
| $\mathbf{M_{17}}$ | Fuzzy MIN_MAX[1] | $E_F$ | 0 | 0 | 1 | 1 | 1 | 2 | 0 | 4 | 0 | [65] |
| $\mathbf{M_{18}}$ | Fuzzy PROMETHEE I | $P_{1F}$ | 1 | 2 | 2 | 1 | 3 | 3 | 3 | 3 | 1 | [66] |
| $\mathbf{M_{19}}$ | Fuzzy PROMETHEE II | $P_{2F}$ | 1 | 2 | 2 | 1 | 3 | 3 | 3 | 3 | 2 | [66] |
| $\mathbf{M_{20}}$ | Fuzzy SAW | $S_F$ | 1 | 2 | 2 | 1 | 1 | 3 | 0 | 3 | 2 | [67] |
| $\mathbf{M_{21}}$ | Fuzzy TOPSIS | $T_F$ | 1 | 2 | 2 | 1 | 1 | 3 | 0 | 3 | 2 | [53] |
| $\mathbf{M_{22}}$ | Fuzzy VIKOR | $V_F$ | 1 | 2 | 2 | 1 | 1 | 3 | 0 | 3 | 2 | [68] |
| $\mathbf{M_{23}}$ | Goal Programming | $G_P$ | 0 | 0 | 2 | 0 | 0 | 0 | 0 | 1 | 0 | [69] |
| $\mathbf{M_{24}}$ | IDRA | $I_D$ | 1 | 2 | 2 | 0 | 0 | 0 | 0 | 3 | 1 | [13] |
| $\mathbf{M_{25}}$ | Lexicographic method | $L_M$ | 1 | 1 | 1 | 0 | 0 | 0 | 0 | 1 | 0 | [70] |
| $\mathbf{M_{26}}$ | MACBETH | $M_B$ | 1 | 3 | 3 | 0 | 0 | 0 | 0 | 3 | 2 | [71] |
| $\mathbf{M_{27}}$ | MAPPAC | $M_P$ | 1 | 2 | 2 | 0 | 0 | 0 | 0 | 3 | 1 | [72] |
| $\mathbf{M_{28}}$ | MAUT | $M_U$ | 1 | 2 | 2 | 0 | 0 | 0 | 0 | 3 | 2 | [73] |
| $\mathbf{M_{29}}$ | MAVT | $M_V$ | 1 | 2 | 2 | 0 | 0 | 0 | 0 | 3 | 2 | [73] |
| $\mathbf{M_{30}}$ | Maximax | $M_X$ | 0 | 0 | 1 | 0 | 0 | 0 | 0 | 1 | 0 | [74] |
| $\mathbf{M_{31}}$ | Maximin | $M_N$ | 0 | 0 | 1 | 0 | 0 | 0 | 0 | 1 | 0 | [74] |
| $\mathbf{M_{32}}$ | Maximin fuzzy method | $M_F$ | 1 | 2 | 2 | 1 | 1 | 2 | 0 | 1 | 0 | [54] |
| $\mathbf{M_{33}}$ | MELCHIOR | $M_C$ | 1 | 1 | 2 | 1 | 2 | 0 | 3 | 3 | 1 | [75] |
| $\mathbf{M_{34}}$ | MIN_MAX[1] | $E_M$ | 0 | 0 | 1 | 0 | 0 | 0 | 0 | 1 | 0 | [74] |
| $\mathbf{M_{35}}$ | NAIADE I | $N_1$ | 0 | 0 | 2 | 1 | 1 | 2 | 0 | 3 | 1 | [76] |
| $\mathbf{M_{36}}$ | NAIADE II | $N_2$ | 0 | 0 | 2 | 1 | 1 | 2 | 0 | 3 | 2 | [76] |
| $\mathbf{M_{37}}$ | ORESTE | $O_R$ | 1 | 1 | 2 | 1 | 2 | 0 | 1 | 3 | 1 | [77] |
| $\mathbf{M_{38}}$ | PACMAN | $P_C$ | 1 | 2 | 2 | 0 | 0 | 0 | 0 | 3 | 1 | [78] |
| $\mathbf{M_{39}}$ | PAMSSEM I | $P_{A1}$ | 1 | 2 | 2 | 1 | 3 | 2 | 3 | 3 | 1 | [79] |
| $\mathbf{M_{40}}$ | PAMSSEM II | $P_{A2}$ | 1 | 2 | 2 | 1 | 3 | 2 | 3 | 3 | 2 | [79] |
| $\mathbf{M_{41}}$ | PRAGMA | $P_G$ | 1 | 2 | 2 | 0 | 0 | 0 | 0 | 3 | 1 | [80] |
| $\mathbf{M_{42}}$ | PROMETHEE I | $P_1$ | 1 | 2 | 2 | 1 | 2 | 0 | 3 | 3 | 1 | [81] |
| $\mathbf{M_{43}}$ | PROMETHEE II | $P_2$ | 1 | 2 | 2 | 1 | 2 | 0 | 3 | 3 | 2 | [81] |
| $\mathbf{M_{44}}$ | QUALIFLEX‘ | $Q_F$ | 1 | 1 | 1 | 0 | 0 | 0 | 0 | 3 | 1 | [82] |
| $\mathbf{M_{45}}$ | REGIME | $R_G$ | 1 | 1 | 1 | 0 | 0 | 0 | 0 | 3 | 1 | [83] |
| $\mathbf{M_{46}}$ | SAW | $S_A$ | 1 | 2 | 2 | 0 | 0 | 0 | 0 | 3 | 2 | [74] |
| $\mathbf{M_{47}}$ | SMART | $S_M$ | 1 | 2 | 2 | 0 | 0 | 0 | 0 | 3 | 2 | [84] |
| $\mathbf{M_{48}}$ | TACTIC | $T_C$ | 1 | 2 | 2 | 1 | 2 | 0 | 1 | 1 | 0 | [15] |
| $\mathbf{M_{49}}$ | TOPSIS | $T_P$ | 1 | 2 | 2 | 0 | 0 | 0 | 0 | 3 | 2 | [85] |
| $\mathbf{M_{50}}$ | UTA | $U_T$ | 1 | 2 | 2 | 0 | 0 | 0 | 0 | 3 | 2 | [86] |
| $\mathbf{M_{51}}$ | VIKOR | $V_K$ | 1 | 2 | 2 | 0 | 0 | 0 | 0 | 3 | 2 | [87] |
| $\mathbf{M_{52}}$ | AHP + fuzzy TOPSIS | $A_H + T_F$ | 1 | 3 | 2 | 1 | 1 | 2 | 0 | 3 | 2 | [56] |
| $\mathbf{M_{53}}$ | Fuzzy AHP + TOPSIS | $A_F + T_P$ | 1 | 3 | 2 | 1 | 1 | 1 | 0 | 3 | 2 | [56] |
| $\mathbf{M_{54}}$ | AHP + VIKOR | $A_H + V_K$ | 1 | 3 | 2 | 0 | 0 | 0 | 0 | 3 | 2 | [88] |
| $\mathbf{M_{55}}$ | DEMATEL | $D_M$ | 1 | 3 | 3 | 0 | 0 | 0 | 0 | 3 | 2 | [89] |
| $\mathbf{M_{56}}$ | REMBRANDT | $R_M$ | 1 | 3 | 3 | 0 | 0 | 0 | 0 | 3 | 2 | [90] |

MIN_MAX[1] - Methods of extracting the minimum and maximum values of the attribute.

c3.1.1 – if the uncertainty concerns the data, does it refer to the weights of criteria or to the variants' criterial performance; possible values are: 1 – criteria, 2 – variants, 3 – both;

c3.1.2 – if the uncertainty concerns the DM's preferences, what thresholds will be used in the decision problem; possible values are: 1 – indifference, 2 – preference, 3 – both.

### 3.2.2. *MCDA methods' properties*

As it was noted above, the descriptors *c* correspond to the characteristics *m*. In such a manner, the considered descriptors were encoded for all considered 56 MCDA methods. Table 2 provides a full description of the MCDA methods depending on all the indicated characteristics (0 means lack of ability). It is worth noting that the inclusion of characteristics relating to all levels of the hierarchy allows to divide MCDA methods into relatively few groups.

### 3.2.3. *Practical mapping between decision problem descriptors and MCDA methods' properties*

The relationships between the set of the MCDA methods' characteristics and the set of a decision problem's descriptors can be presented by analyzing an exemplary decision problem and the procedure of the MCDA method selection for solving it. In [91], a decision-making problem of constructing a ranking of premises for urban distribution centers was considered. It considered three alternative locations in terms of 11 criteria. During the selection of the MCDA method for the given decision-making problem, a full set of descriptors was used, i.e. $\tilde{c} = c$. The decision-making

**Table 3**
The rules of selecting a suitable MCDA method.

| MCDA method properties | | $m_{i1}$ | $m_{i2}$ | $m_{i3}$ | $m_{i4}$ | $m_{i1.1}$ | $m_{i3.1}$ | $m_{i4.1}$ | $m_{i3.1.1}$ | $m_{i3.1.2}$ | Subset of MCDA methods | |
|---|---|---|---|---|---|---|---|---|---|---|---|---|
| | | | | | | | | | | | Names | Abbreviations |
| Rules | $R_1$ | 0 | 1 | 0 | 1 | 0 | 0 | 0 | 0 | 0 | Maximax, Maximin, MIN_MAX[1] | $\{M_X, M_N, E_M\}$ |
| | $R_2$ | 0 | 1 | 1 | 4 | 0 | 1 | 0 | 2 | 0 | FuzzyMIN_MAX[1] | $\{E_F\}$ |
| | $R_3$ | 0 | 1 | 1 | 3 | 0 | 2 | 1 | 0 | 3 | ELECTRE IV | $\{E_4\}$ |
| | $R_4$ | 0 | 2 | 0 | 1 | 0 | 0 | 0 | 0 | 0 | Goal Programming | $\{G_P\}$ |
| | $R_5$ | 0 | 2 | 1 | 3 | 0 | 1 | 1 | 2 | 0 | NAIADE I | $\{N_1\}$ |
| | $R_6$ | 0 | 2 | 1 | 3 | 0 | 1 | 2 | 2 | 0 | COMET, NAIADE II | $\{C_T, N_2\}$ |
| | $R_7$ | 1 | 1 | 0 | 1 | 1 | 0 | 0 | 0 | 0 | ARGUS, Lexicographic method | $\{A_G, L_M\}$ |
| | $R_{11}$ | 1 | 1 | 0 | 1 | 2 | 0 | 0 | 0 | 0 | ELECTRE I | $\{E_1\}$ |
| | $R_8$ | 1 | 1 | 0 | 3 | 1 | 0 | 1 | 0 | 0 | QUALIFLEX, REGIME | $\{Q_F, R_G\}$ |
| | $R_{12}$ | 1 | 1 | 0 | 3 | 2 | 0 | 1 | 0 | 0 | ELECTRE II | $\{E_2\}$ |
| | $R_{13}$ | 1 | 2 | 0 | 3 | 2 | 0 | 1 | 0 | 0 | IDRA, MAPPAC, PACMAN, PRAGMA | $\{I_D, M_P, P_C, P_G\}$ |
| | $R_{14}$ | 1 | 2 | 0 | 3 | 2 | 0 | 2 | 0 | 0 | EVAMIX, MAUT, MAVT, SAW, SMART, TOPSIS, UTA, VIKOR | $\{E_V, M_U, M_V, S_A, S_M, T_P, U_T, V_K\}$ |
| | $R_{26}$ | 1 | 2 | 0 | 3 | 3 | 0 | 2 | 0 | 0 | AHP + TOPSIS, AHP + VIKOR | $\{A_H + T_P, A_H + V_K\}$ |
| | $R_{15}$ | 1 | 2 | 1 | 1 | 2 | 1 | 0 | 2 | 0 | Maximin fuzzy method | $\{M_F\}$ |
| | $R_{17}$ | 1 | 2 | 1 | 1 | 2 | 2 | 0 | 0 | 1 | TACTIC | $\{T_C\}$ |
| | $R_{18}$ | 1 | 2 | 1 | 1 | 2 | 2 | 0 | 0 | 3 | ELECTRE IS | $\{E_S\}$ |
| | $R_{19}$ | 1 | 2 | 1 | 2 | 2 | 2 | 0 | 0 | 3 | ELECTRE TRI | $\{E_T\}$ |
| | $R_9$ | 1 | 2 | 1 | 3 | 1 | 2 | 1 | 0 | 1 | ORESTE | $\{O_R\}$ |
| | $R_{10}$ | 1 | 2 | 1 | 3 | 1 | 2 | 1 | 0 | 3 | MELCHIOR | $\{M_C\}$ |
| | $R_{16}$ | 1 | 2 | 1 | 3 | 2 | 1 | 2 | 3 | 0 | Fuzzy SAW, Fuzzy TOPSIS, Fuzzy VIKOR | $\{S_F, T_F, V_F\}$ |
| | $R_{20}$ | 1 | 2 | 1 | 3 | 2 | 2 | 1 | 0 | 3 | ELECTRE III, PROMETHEE I | $\{E_3, P_1\}$ |
| | $R_{21}$ | 1 | 2 | 1 | 3 | 2 | 2 | 2 | 0 | 3 | PROMETHEE II | $\{P_2\}$ |
| | $R_{22}$ | 1 | 2 | 1 | 3 | 2 | 3 | 1 | 2 | 3 | PAMSSEM I | $\{P_{A1}\}$ |
| | $R_{24}$ | 1 | 2 | 1 | 3 | 2 | 3 | 1 | 3 | 3 | Fuzzy PROMETHEE I | $\{P_{1F}\}$ |
| | $R_{23}$ | 1 | 2 | 1 | 3 | 2 | 3 | 2 | 2 | 3 | PAMSSEM II | $\{P_{A2}\}$ |
| | $R_{25}$ | 1 | 2 | 1 | 3 | 2 | 3 | 2 | 3 | 3 | Fuzzy PROMETHEE II | $\{P_{2F}\}$ |
| | $R_{27}$ | 1 | 2 | 1 | 3 | 3 | 1 | 2 | 1 | 0 | Fuzzy AHP + TOPSIS | $\{A_F + T_P\}$ |
| | $R_{28}$ | 1 | 2 | 1 | 3 | 3 | 1 | 2 | 2 | 0 | AHP + fuzzy TOPSIS | $\{A_H + T_F\}$ |
| | $R_{29}$ | 1 | 2 | 1 | 3 | 3 | 1 | 2 | 3 | 0 | Fuzzy AHP + fuzzy TOPSIS, Fuzzy ANP + fuzzy TOPSIS | $\{A_F + T_F, A_{NF} + T_F\}$ |
| | $R_{30}$ | 1 | 3 | 0 | 3 | 3 | 0 | 2 | 0 | 0 | AHP, ANP, MACBETH, DEMATEL, REMBRANDT | $\{A_H, A_N, M_B, D_M, R_M\}$ |
| | $R_{31}$ | 1 | 3 | 1 | 3 | 3 | 1 | 2 | 3 | 0 | Fuzzy AHP, Fuzzy ANP | $\{A_F, A_{NF}\}$ |

MIN_MAX[1] - methods of extracting the minimum and maximum values of the attribute.

problem includes the weights of criteria in quantitative form, so the descriptors of the decision problem have taken the values $c_1 = 1$, $c_{1.1} = 2$. In addition, the efficiency of the variants was expressed on a quantitative scale ($c_2 = 2$). The decision-making problem was characterized by uncertainty ($c_3 = 1$), where the uncertainty referred to the input data ($c_{3.1} = 1$), and in particular to the weightings of the criteria and performance of the decision variants ($c_{3.1.1} = 3$, $c_{3.1.2} = 0$). The considered decision problematic was the problematic of ranking, and the obtained solution was a complete ranking, i.e. ranking without incomparability ($c_4 = 3$, $c_{4.1} = 2$). It is easy to notice that the individual descriptors $c$ correspond to the $m$ characteristics of the same values, i.e. $m_{i1} = 1$, $m_{i1.1} = 2$, $m_{i2} = 2$, $m_{i3} = 1$, $m_{i3.1} = 1$, $m_{i3.1.1} = 3$, $m_{i3.1.2} = 0$, $m_{i3.4} = 3$, $m_{i4.1} = 2$ etc. Analysis of Table 2 allows to notice three MCDA methods having such characteristics vectors: Fuzzy SAW($M_{20}$), Fuzzy TOPSIS ($M_{21}$) and Fuzzy VIKOR ($M_{22}$).

#### 3.2.4. MCDA method properties' explanation

When we have a given decision problem, its requirements with relations to properties of individual MCDA methods can be determined.

The property $m_1$ refers to the weights of the criteria. MCDA methods may use qualitative, quantitative or relative weights, as well as may not use criteria weights. For example, in [92], the criteria weights are not used, which results in the properties related to the criteria $m_1$ and $m_{1.1}$ not being met. On the other hand, in [93], quantitative weights of criteria were applied, which means that the property $m_1$ is met, and the property $m_{1.1}$ obtained the value of 2. Finally, in [94], the weights of criteria were compared pairwise in the form of a comparison matrix, thus providing a weights vector. Therefore, the property $m_1$ was met, and the property $m_{1.1}$ obtained the value of 3.

The second property describes the scale at which the performance of the variants in each of the criteria are compared or determined. As in the case of the criteria weights, this scale can be qualitative, quantitative or relative. In [89,95], only the significance of individual criteria related to Green Supply Chain Management (GSCM) was examined, without considering any decision variants. This means that the property $m_2$ of the decision problem is not met. In contrast, property $m_2$ is met for example in [96], where the variants were compared on a qualitative scale and, therefore, property $m_2$ is given the value of 1. In [69], a quantitative scale was used to compare the variants, so that $m_2$ obtained the value of 2. Eventually, in [97], the comparative scale was used for comparisons of variants (pairwise comparison matrix), therefore, property $m_2$ obtains the value of 3.

Property $m_3$ refers to the uncertainty of the decision problem. The uncertainty may refer to the input data describing the criteria weights or the variants' performance in each criterion. In such case, the data is expressed with the use of fuzzy numbers. On the other hand, the uncertainty may also apply to the preferences of the decision makers. This kind of uncertainty is expressed with the use of the thresholds of indifference and preference. The indifference threshold determines the difference in the criterion performance of individual variants, at which they can be considered to be equally good. On the other hand, the threshold of preference defines the difference in the performance of the variants, in which one of the variants is considered to be definitely better than the other. Uncertainty is included e.g. in [98]. It is an uncertainty related to data at the level of the criteria weights and the performance of the criteria. Therefore, the properties $m_3$, $m_{3.1}$ and $m_{3.1.1}$ are fulfilled, with $m_{3.1}$ being 1 and $m_{3.1.1}$ being 3. In contrast, in [99], uncertainty about the decision maker's preferences occur, so the thresholds of indifference and preference were applied. Therefore, the properties $m_3$, $m_{3.1}$ and $m_{3.1.2}$ are met in this case, with $m_{3.1}$ being 2 and $m_{3.1.2}$ being 3.

Last, but not least, the $m_4$ property refers to the decision problematics. It should be clarified that the methods dealing with the ranking problem, also allow to solve the choice problem, and, therefore, one of the possible values of the property $m_4$ includes both the ranking and the choice problems. If the MCDA method considers a ranking problem, it may provide the results in the form of a full (total order) or partial ranking (partial order). A method supporting the total order usually allows to obtain global performance values of the variants in numerical form and to determine for each pair of variants which one is better. In contrast, methods supporting the partial order do not provide a full comparability of the variants and most often express the global efficiency of variants on an ordinal scale, which, additionally, does not allow indicate which variant is better for any pair of variants. For example, in [96], the issue of choice is considered, so the property $m_4$ takes the value of 1. The problem of ranking is considered e.g. in [100] and [101], where the property $m_4$ obtains the value of 3. In the former, a total order of variants is produced, thus the property $m_{4.1}$ obtains the value of 2, whereas in the latter a partial order of variants was obtained, so the property $m_{4.1}$ takes the value of 1.

### 3.3. Formal representation of the MCDA methods' properties

When looking for a formal approach for selecting an MCDA method, applying a classifier, as it was done in a number of studies [21,22,32], seems to be an interesting concept. In current work we propose the set of descriptors identifying MCDA methods' properties for a particular decision situation, presented below:

The descriptor $c_1$ checks if weights of any kind will be used in the decision problem:

$$c_1(DP) = \begin{cases} 1 & (\exists\{g_1 \succ g_2 \succ \ldots \succ g_m\}) \otimes \left(\underset{g_i,g_j \in G}{\forall} \exists r : \left|g_i - g_j\right| = r\right) \\ & \otimes \left(\underset{g_i,g_j \in G}{\forall} \exists W_{|G|\times|G|} : w_{ij} = g_i/g_j\right) \\ 0 & otherwise \end{cases}$$

where:

- $r$ – the quantitative difference between the weights of criteria ($g_i$ and $g_j$),
- $w_{ij}$ – a relative weight of a criterion $g_i$ in respect to a criterion $g_j$,
- $W_{|G|x|G|}$ – matrix of pairwise comparisons, where size is equal to the size of a set of criteria $G$.

The descriptor $c_{1.1}$ is responsible for distinguishing the type of weights used, which can be expressed on the scale of one of the following types: qualitative, quantitative or relative:

$$c_{1.1}(DP) = \begin{cases} 1 & \exists\{g_1 \succ g_2 \succ \ldots \succ g_m\} \\ 2 & \underset{g_i,g_j \in G}{\forall} \exists r : \left|g_i - g_j\right| = r \\ 3 & \underset{g_i,g_j \in G}{\forall} \exists W_{|G|\times|G|} : w_{ij} = g_i/g_j \end{cases}$$

The descriptor $c_2$ distinguishes the type of scale on which the decision variants will be compared. These can be: qualitative, quantitative or relative scales. The descriptor $c_2$ also includes a situation when the variants are not compared. It is due to the fact that in situations where MCDA methods are used, such situations still may occur, even though all of the considered MCDA methods include comparisons of variants. The descriptor $c_2$ value is determined according to formula:

$$c_2(DP) = \begin{cases} 1 & \forall_{\substack{a \in A \\ g_i \in G}} \exists\{g_i(a_1) \succ g_i(a_2) \succ \ldots \succ g_i(a_m)\} \\ 2 & \forall_{\substack{a_j,a_k \in A \\ g_i \in G}} \exists r : \left|g_i(a_j) - g_i(a_k)\right| = r \\ 3 & \forall_{\substack{a_j,a_k \in A \\ g_i \in G}} \exists E_{|A|\times|A|} : e_{jk} = g_i(a_j)/g_i(a_k) \\ 0 & otherwise \end{cases}$$

where:

$g_i(a)$ – the performance of a variant $a$ concerning a criterion $i$ and a weight $g_i$,
$r$ – the quantitative difference between performances of variants $a$ with respect to criterion $g$,
$e_{jk}$ – relative criterion performance (for the criterion $g$) of variant $a_j$ with respect to a variant $a_k$,
$E_{|A|x|A|}$ – matrix of pairwise comparisons, where size is equal to the cardinality set of variants $A$.

The descriptor $c_3$ checks if there is data or preference uncertainty in the decision problem:

$$c_3(DP) = \begin{cases} 1 & \left(\underset{g_i \in G}{\forall} \exists \tilde{N}_{fuzzy} = (n_l; n_u; \alpha_F; \beta_F)_{LR} : g_i = \tilde{N}_{fuzzy}\right) \vee \left(\forall_{\substack{a_j \in A \\ g_i \in G}} \exists \tilde{N}_{fuzzy} = (n_l; n_u; \alpha_F; \beta_F)_{LR} : g_i(a_j) = \tilde{N}_{fuzzy}\right) \vee \ldots \\ & \left(\forall_{\substack{a_j, a_k \in A \wedge j \neq k \\ g_i \in G}} \exists (q) : \begin{cases} g_i(a_j) - g_i(a_k) \le q \Rightarrow g_i(a_j) \sim g_i(a_k) \\ g_i(a_j) - g_i(a_k) > q \Rightarrow g_i(a_j) \succ g_i(a_k) \end{cases}\right) \vee \left(\forall_{\substack{a_j, a_k \in A \wedge j \neq k \\ g_i \in G}} \exists (p) : \begin{cases} g_i(a_j) - g_i(a_k) = 0 \Rightarrow g_i(a_j) \sim g_i(a_k) \\ p > g_i(a_j) - g_i(a_k) > 0 \Rightarrow g_i(a_j) \succ_{weak} g_i(a_k) \\ g_i(a_j) - g_i(a_k) \ge p \Rightarrow g_i(a_j) \succ g_i(a_k) \end{cases}\right) \\ 0 & otherwise \end{cases}$$

where:

$N_{fuzzy}$ – triangular of a trapezoidal fuzzy number,
$(n_l, n_u, \alpha_F, \beta_F)$ – parameters of a fuzzy number membership function,
$q$ – indifference threshold,
$p$ – preference threshold.

The descriptor $c_{3.1}$ verifies if the uncertainty is related particularly to input data, to preferences, or to both of them:

$$c_{3.1}(DP) = \begin{cases} 1 & \left(\underset{g_i \in G}{\forall} \exists \tilde{N}_{fuzzy} = (n_l; n_u; \alpha_F; \beta_F)_{LR} : g_i = \tilde{N}_{fuzzy}\right) \vee \left(\forall_{\substack{a_j \in A \\ g_i \in G}} \exists \tilde{N}_{fuzzy} = (n_l; n_u; \alpha_F; \beta_F)_{LR} : g_i(a_j) = \tilde{N}_{fuzzy}\right) \\ 2 & \left(\forall_{\substack{a_j, a_k \in A \wedge j \neq k \\ g_i \in G}} \exists (q) : \begin{cases} g_i(a_j) - g_i(a_k) \le q \Rightarrow g_i(a_j) \sim g_i(a_k) \\ g_i(a_j) - g_i(a_k) > q \Rightarrow g_i(a_j) \succ g_i(a_k) \end{cases}\right) \vee \left(\forall_{\substack{a_j, a_k \in A \wedge j \neq k \\ g_i \in G}} \exists (p) : \begin{cases} g_i(a_j) - g_i(a_k) = 0 \Rightarrow g_i(a_j) \sim g_i(a_k) \\ p > g_i(a_j) - g_i(a_k) > 0 \Rightarrow g_i(a_j) \succ_{weak} g_i(a_k) \\ g_i(a_j) - g_i(a_k) \ge p \Rightarrow g_i(a_j) \succ g_i(a_k) \end{cases}\right) \\ 3 & \left(\underset{g_i \in G}{\forall} \exists \tilde{N}_{fuzzy} = (n_l; n_u; \alpha_F; \beta_F)_{LR} : g_i = \tilde{N}_{fuzzy}\right) \vee \left(\forall_{\substack{a_j \in A \\ g_i \in G}} \exists \tilde{N}_{fuzzy} = (n_l; n_u; \alpha_F; \beta_F)_{LR} : g_i(a_j) = \tilde{N}_{fuzzy}\right) \vee \ldots \\ & \left(\forall_{\substack{a_j, a_k \in A \wedge j \neq k \\ g_i \in G}} \exists (q) : \begin{cases} g_i(a_j) - g_i(a_k) \le q \Rightarrow g_i(a_j) \sim g_i(a_k) \\ g_i(a_j) - g_i(a_k) > q \Rightarrow g_i(a_j) \succ g_i(a_k) \end{cases}\right) \vee \left(\forall_{\substack{a_j, a_k \in A \wedge j \neq k \\ g_i \in G}} \exists (p) : \begin{cases} g_i(a_j) - g_i(a_k) = 0 \Rightarrow g_i(a_j) \sim g_i(a_k) \\ p > g_i(a_j) - g_i(a_k) > 0 \Rightarrow g_i(a_j) \succ_{weak} g_i(a_k) \\ g_i(a_j) - g_i(a_k) \ge p \Rightarrow g_i(a_j) \succ g_i(a_k) \end{cases}\right) \end{cases}$$

The descriptor $c_{3.1.1}$ further divides the input data uncertainty into the ones in which the fuzzy sets were used to the criteria's weights, to the variants' performance or to both of them:

$$c_{3.1.1}(DP) = \begin{cases} 1 & \underset{g_i \in G}{\forall} \exists \tilde{N}_{fuzzy} = (n_l; n_u; \alpha_F; \beta_F)_{LR} : g_i = \tilde{N}_{fuzzy} \\ 2 & \forall_{\substack{a_j \in A \\ g_i \in G}} \exists \tilde{N}_{fuzzy} = (n_l; n_u; \alpha_F; \beta_F)_{LR} : g_i(a_j) = \tilde{N}_{fuzzy} \\ 3 & \forall_{\substack{a_j \in A \\ g_i \in G}} \exists \tilde{N}_{fuzzy} = (n_l; n_u; \alpha_F; \beta_F)_{LR} : g_i, g_i(a_j) = \tilde{N}_{fuzzy} \end{cases}$$

The descriptor $c_{3.1.2}$ further divides the preference uncertainty by distinguishing the situations in which the thresholds of indifference, preference or both of them were used:

$$c_{3.1.2}(DP) = \begin{cases} 1 & \forall_{\substack{a_j, a_k \in A \wedge j \neq k \\ g_i \in G}} \exists (q) : \begin{cases} g_i(a_j) - g_i(a_k) \le q \Rightarrow g_i(a_j) \sim g_i(a_k) \\ g_i(a_j) - g_i(a_k) > q \Rightarrow g_i(a_j) \succ g_i(a_k) \end{cases} \\ 2 & \forall_{\substack{a_j, a_k \in A \wedge j \neq k \\ g_i \in G}} \exists (p) : \begin{cases} g_i(a_j) - g_i(a_k) = 0 \Rightarrow g_i(a_j) \sim g_i(a_k) \\ p > g_i(a_j) - g_i(a_k) > 0 \Rightarrow g_i(a_j) \succ_{weak} g_i(a_k) \\ g_i(a_j) - g_i(a_k) \ge p \Rightarrow g_i(a_j) \succ g_i(a_k) \end{cases} \\ 3 & \forall_{\substack{a_j, a_k \in A \wedge j \neq k \\ g_i \in G}} \exists (p, q) : \begin{cases} g_i(a_j) - g_i(a_k) \le q \Rightarrow g_i(a_j) \sim g_i(a_k) \\ p > g_i(a_j) - g_i(a_k) > q \Rightarrow g_i(a_j) \succ_{weak} g_i(a_k) \\ g_i(a_j) - g_i(a_k) \ge p \Rightarrow g_i(a_j) \succ g_i(a_k) \end{cases} \end{cases}$$

The descriptor $c_4$ checks which problematic is considered in the decision problem:

$$c_4(DP) = \begin{cases} 1 & f\left(\max_{u \in A' \subset A} \left\{S_D(u) : \dim(u) = \min\left\{\dim(v);\ \underset{v \in A \setminus A'}{\forall} \underset{q \in A'}{\exists} (\neg S_D(q))\right\}\right\}\right) \\ 2 & f(u_B);\ \underset{u \in A}{\exists} u_B = u \wedge \underset{v \in A, v \neq u}{\forall} \eta(u) > \eta(v) \\ 3 & f(k_R);\ \underset{k \in KR}{\exists} k_R = k \wedge \underset{kp \in KR, kp \neq k}{\forall} \upsilon(k) \approx \upsilon(kp) \\ 4 & f\left(\max_{u \in A' \subset A} \left\{S_D(u) : \dim(u) = \min\left\{\dim(v);\ \underset{v \in A \setminus A'}{\forall} \underset{q \in A'}{\exists} (\neg S_D(q))\right\}\right\}\right) \vee f(u_B);\ \underset{u \in A}{\exists} u_B = u \wedge \underset{v \in A, v \neq u}{\forall} \eta(u) > \eta(v) \end{cases}$$

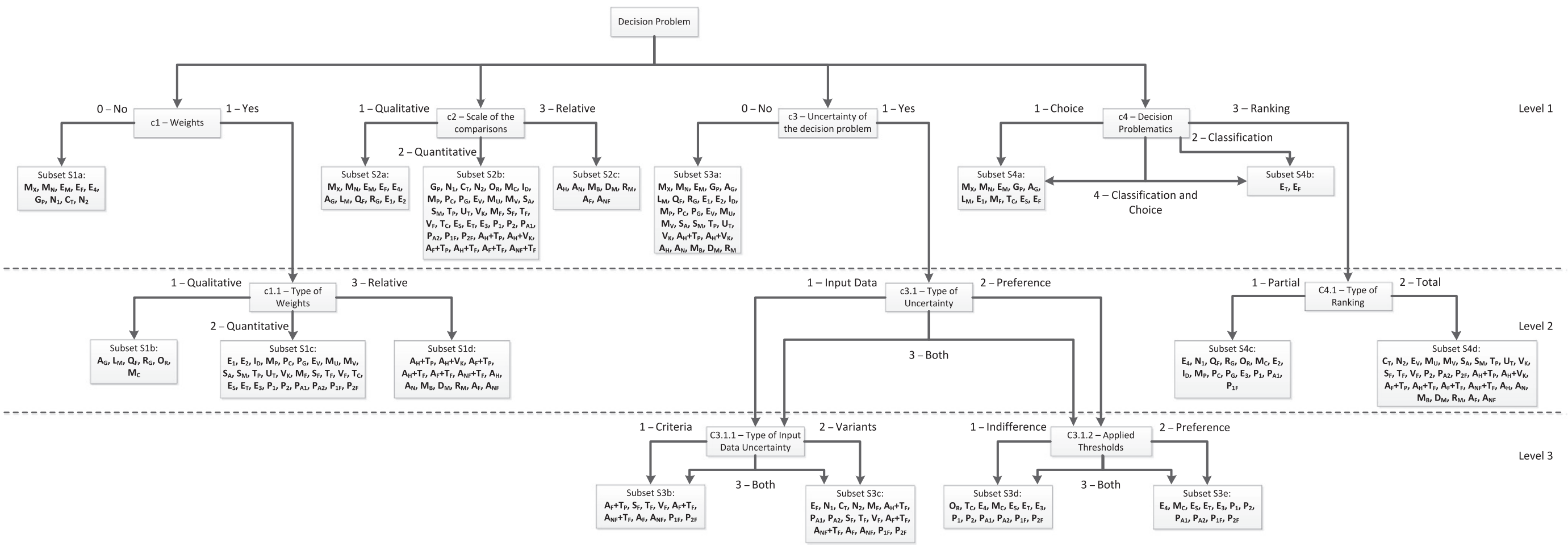

**Fig. 2.** The decision tree of selecting a suitable MCDA method on the basis of the proposed descriptors.

where:

dim – size,
$S_D()$ – DM's satisfaction with the variant,
$\eta$ – the norm related to the certain values,
$KR$ – the set of equivalence classes of variants from the set $A$,
$\approx$ – the relation of partial or complete order.

If the problematic of ranking is considered, the descriptor $c_{4.1}$ checks whether a partial or full order of variants is expected:

$$c_{4.1}(DP) = \begin{cases} 1 & \underset{E=G(A)}{\forall} \ \exists a_i, a_j : E(a_i)\ R\ E(a_j) \\ 2 & \underset{E=G(A)}{\forall} \ \exists \{E(a_1) \succ E(a_2) \succ \ldots \succ E(a_m)\} \end{cases}$$

where:

$E(a)$ – a global performance of variant $a$,
$R$ – incomparability relation.

### 3.4. Presentation of the MCDA methods' properties using tree representation

The applied classifier can also be presented in a form of decision trees and, in consequence, the whole classification process would have the form of a method selection tree, as it is called by Guitouni and Martel [20]. The decision tree of selecting a suitable MCDA method on the basis of the proposed descriptors is presented on Fig. 2.

The tree presents the problem of an MCDA method selection depending on the information about the decision problem known to the DM. To specify a subset of methods that meet the descriptors describing the decision problem, the algebra of sets should be used.

If the DM has full information about the decision problem, and, therefore, knows what kind of weights should be used, on what scale the variants should be compared, what kind of uncertainty should be included in the decision problem and what the decision problematic is, then the appropriate subset of MCDA methods is determined as intersection of relevant subsets *S1a: S4d* (horizontal approach). For example, if the decision problem is described by descriptors $c_1=1$, $c_{1.1}=2$, $c_2=3$, $c_3=0$ and $c_4=1$, the subset of methods is the intersection of the sets $S1c \cap S2b \cap S3a \cap S4a$. The descriptors $c_{3.1.1}$ and $c_{3.1.2}$ taking the value of 3 and the descriptor $c_4$ taking the value of 4 are special cases. In particular, regarding the descriptor $c_{3.1.1}=3$, to account for the MCDA methods considering the data uncertainty for both weights of criteria and the variants, the intersection of sets $S3b \cap S3c$ should be used. In the other cases mentioned above, the intersection of sets should be used analogously.

On the other hand, for the decision problem which the DM cannot fully define, it may be necessary to apply the union of sets (vertical approach). This allows the inclusion of more general descriptors, occurring at levels 1 and 2 of the hierarchy, to which no subsets of methods have been directly assigned. For example, if weights are included in the decision problem, but the scale on which they should be expressed is unknown, the determination of the appropriate methods takes place using the union of the sets $S1b \cup S1c \cup S1d$.

Consequently, the adaptation of the MCDA method to an incompletely defined decision problem is a combination of the vertical approach, referring to the lack of information about the decision-making problem, and the horizontal approach, referring to the certain information. For example, if the DM knowns that in the decision problem: quantitative weights should be used ($c_1=1$ and $c_{1.1}=2$), there exists data uncertainty related both to the weights of criteria and to the variants ($c_3=1$, $c_{3.1}=1$, $c_{3.1.1}=3$), the ranking problematic should be considered, and a full order should be obtained ($c_4=3$ and $c_{4.1}=2$), but there is uncertainty on what scale the variants should be compared, the subset of the appropriate methods will result from the $S1c \cap S3b \cap S3c \cap S4d \cap (S2a \cup S2b \cup S2c)$ operation.

### 3.5. Rules database generation

It needs to be noted that the characteristics describing properties of individual multi-criteria methods would also be used as conditional attributes, whereas specific MCDA methods would be decision attributes which constitutes foundations for the MCDA selection rules set. Of the 56 MCDA methods considered, there were only 31 unique sequences of encoded characteristics. This implicates that some of the methods have identical characteristics. A subset of suitable MCDA methods can be recommended. The characteristics of the problem are not reproducible as it is in the case of the MCDA methods. The problem characteristics should be identified each time separately. On the basis of the proposed properties the expert rule base is obtained. The set of rules is presented in Table 3. In addition, rules at various levels of the hierarchy are included here. The first level, limited to the most general properties ($m_1$, $m_2$, $m_3$, $m_4$), allows defining 13 distinct rules. On the second level of the hierarchy, which includes more specific properties of MCDA methods, a total of 25 rules was distinguished, allowing the selection of MCDA methods depending on the properties of the decision problem. Last, but not least, the third and most detailed level of the hierarchy allows to define the aforementioned 31 rules.

The presented table shows that the MCDA method can be selected depending on the values of the descriptors at a successive level. However it is clearly visible that the lack of knowledge about a particular level decreases the quality of recommendation. A detailed analysis is presented in Supplementary material – Section 4.When analyzing the number of methods assigned to the decision problem by individual rules, it is easy to notice that the $R_{14}$ rule is the most capacious one. Based on this rule, eight methods are indicated as appropriate to solve a problem of a specific character: EVAMIX, MAUT, MAVT, SAW, SMART, TOPSIS, UTA, VIKOR. The high number of methods in this rule results from the great similarity of the majority of the methods included in it. The MAVT method is basically a simplification of the MAUT method, with the only significant difference being the fact that during the aggregation MAVT uses a value function, and MAUT – a utility function. The value function, in contrast to the utility function, does not take into account the risk (probability) [102]. The SAW method, on the other hand, is the simplest case of the MAVT method, where the additive value function is normalized to the [0,1] interval [103]. Similarly, the SMART method is a simplification of MAVT / MAUT, in which an additive model is used with a linear approximation of the utility / value function [104]. In turn, the UTA method uses partial, criterial usability / values functions that are monotonic and sectionally linear. The partial functions are then aggregated using the additive value function [38]. It can be, therefore, concluded that the MAVT, SAW, SMART and UTA methods are special cases of the MAUT method. Equally significant similarity can be observed between the TOPSIS and VIKOR methods, which are based on the same principles. They differ only in the aggregation function and the normalization method used. In TOPSIS, the aggregation function minimizes the distance to the ideal solution and maximizes the distance from the anti-ideal solution, whereas in VIKOR, the aggregation function only minimizes the distance to the ideal solution. As for normalization, vector normalization is used in TOPSIS and linear normalization is used in VIKOR [87]. Based on the aforementioned observations, it can be therefore concluded that the $R_{14}$ rule essentially includes three different subsets of methods similar to each other.

It needs to be noted that the rule-based approach presented in Table 3 is different than the approach based on the algebra of sets and the tree structure (Fig. 2) presented in Section 3.4. In the tree structure, descriptors $c_{3.1}$, $c_{3.1.1}$, $c_{3.1.2}$ and $c_4$ assign an MCDA method to one of two disjoint sets or to the intersection of the sets. In turn, in the rules presented in Table 3, each of the possibilities (both subsets, as well as their intersection) is coded separately.

### 3.6. Uncertainty handling in the decision problem description

Nevertheless, it needs to be noted that the aforementioned analysis can be oversimplifying of the real decision making problem. Often, the DM might not have the full knowledge of the decision problem, or the possessed knowledge might not provide the full knowledge of any of the levels of hierarchy, thus introducing uncertainty to the decision-making. Therefore, the DM can know the values of as much as 4, 7 or 9 classifiers for a single, two or three levels of descriptors respectively, or as little as a single classifier.

For this reason, the authors decided to introduce a more profound modelling of the uncertainty of data and the decision-making problem. In the next step of the empirical research, a set of all possible 450,000 rules for all possible values of each classifier was generated. However, in majority of the rules, some of the classifiers were in conflict – e.g. $c_1 = 0$ (no weights) but $c_{1.1} = 3$ (relative weights) – and, therefore, a subset of the rules was extracted, based on the criteria presented in Table S5 in Supplementary material – Section 5. The extraction process consisted of four steps, depicted on Fig. S5 in Supplementary material – Section 5. In the first step, the full set of rules was filtered four times to extract the rules with the valid values of $c_1$ and $c_{1.1}$ classifiers. In the second step, the output of the first step was then filtered with the valid values of the $c_2$ classifier. The output was then filtered with the valid values of the $c_3$, $c_{3.1}$, $c_{3.1.1}$ and $c_{3.1.2}$ classifiers in step three. Eventually, the output was filtered with the valid values of the classifiers $c_4$ and $c_{4.1}$. As a result, the original set of 450 thousand rules was reduced to 4,536 rules. Furthermore, after the removal of the rules returning 0 methods, a final set of 656 rules was obtained. A similar procedure was then performed for the hierarchies consisting of two levels and a single level of classifiers.

Fig. 3 illustrates a histogram-like analysis of the possible MCDA selection rules, based on the number of unknown MCDA characteristics, as well as the complexity of the structure. The x-axis and right y-axis represent the number of the characteristics that are unknown, whereas the left y-axis represents the growth in number of methods that possibly meet these unknown characteristics. The bars in the chart represent the precise count of methods matching every single rule, whilst the dashed line represents their average count for the corresponding number of unknowns. The presented rules are limited to the 656 ones that returned at least one method. A detailed analysis of the rule sets containing all rules, including the ones returning empty sets of methods, is provided in Supplementary material – Section 6.

The analysis of Fig. 3 and Table 4 allows to observe that in case of the 1-level hierarchy of the decision rules, if the DM cannot decide on a single characteristic, on average, the number of matching MCDA methods would be almost 2 times higher than in case of a single unknown in the 2-level hierarchy. On the other hand, the difference in case of the 2 levels and 3 levels of characteristics is equal to only 0.8046, for a single unknown. In case of two unknown values of the MCDA characteristics, the average number of possible matching methods is over 2.5 times higher than in case of 2 levels and almost 4 times higher than in case of 3 levels. In order to match the number of methods produced by a single unknown in the 1-level rule set, at least 3 variables should be unknown for the 2-level one and 5 for the 3-level one.

When Fig. 3c is analyzed in detail, it can be observed that when a single characteristic is unknown to the decision maker, the 3-level decision framework still allows to limit the matching number of MCDA methods to a range of 1 to 12, with average value equal to 2.1446. If the number of unknown decision problem descriptors grows to two values, a significant increase of non-empty rules can be observed, to a total of 131. The average number of matched methods increases only slightly to 2.5191. A similar growth of possibly matching methods can be observed also when the count of unknown abilities grows to 3 and 4, with the average equal to 3.0511 and 3.7719 respectively. However, if the number of unknown characteristics increases any further, the speed of growth of the number of methods starts to increase rapidly which fact is illustrated on Fig. 4a.

The average count of matching methods for cases when 5, 6, 7 or 8 of the total of 9 characteristics are unknown is equal to 5.2099, 8.0400, 14.4545 and 29 respectively. This growth can be mapped by a 4-degree polynomial function with the $R^2$ equal to 0.9997. It is also important to note from Fig. 3c, that along with the increase of the count of methods, the number of rules decreases when more MCDA characteristics are unknown.

Fig. 3a, 3b depict the two remaining scenarios when the knowledge about the decision problem is structured only into two levels (Fig. 3b) or into a sequence of 4 main classifiers (Fig. 3a). A linear

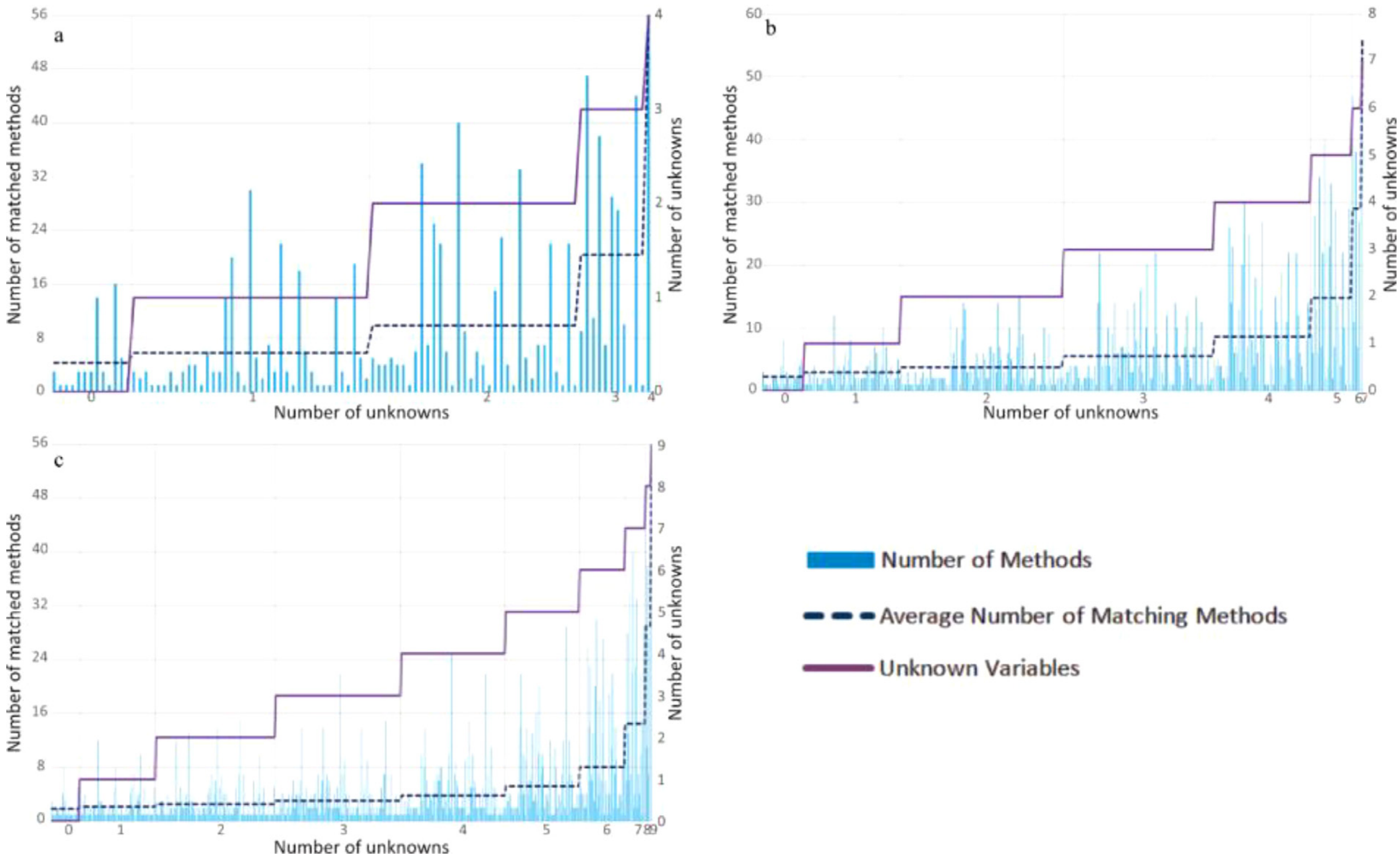

**Fig. 3.** A histogram-like analysis of the possible MCDA selection rules depending on the number of unknown characteristics, in cases of a single level (a), two levels (b) and three levels (c) of MCDA methods' properties, excluding the rules returning 0 methods.

**Table 4**
Comparison of the minimum, average and maximum number of methods for classifiers organized into one, two or three levels.

| Unknowns | 1 Level | | | 2 Levels | | | 3 Levels | | |
|---|---|---|---|---|---|---|---|---|---|
| | Min | Mean | Max | Min | Mean | Max | Min | Mean | Max |
| 0 | 1 | 4.3077 | 16 | 1 | 2.2400 | 8 | 1 | 1.8065 | 8 |
| 1 | 1 | 5.7436 | 30 | 1 | 2.9492 | 12 | 1 | 2.1446 | 12 |
| 2 | 1 | 9.8824 | 40 | 1 | 3.7374 | 15 | 1 | 2.5191 | 15 |
| 3 | 1 | 20.3636 | 47 | 1 | 5.5000 | 22 | 1 | 3.0511 | 22 |
| 4 | 56 | 56.0000 | 56 | 1 | 8.5763 | 30 | 1 | 3.7719 | 25 |
| 5 | | | | 1 | 14.8000 | 40 | 1 | 5.2099 | 29 |
| 6 | | | | 7 | 29.0000 | 47 | 1 | 8.0400 | 30 |
| 7 | | | | 56 | 56.0000 | 56 | 1 | 14.4545 | 40 |
| 8 | | | | | | | 7 | 29.0000 | 47 |
| 9 | | | | | | | 56 | 56.0000 | 56 |

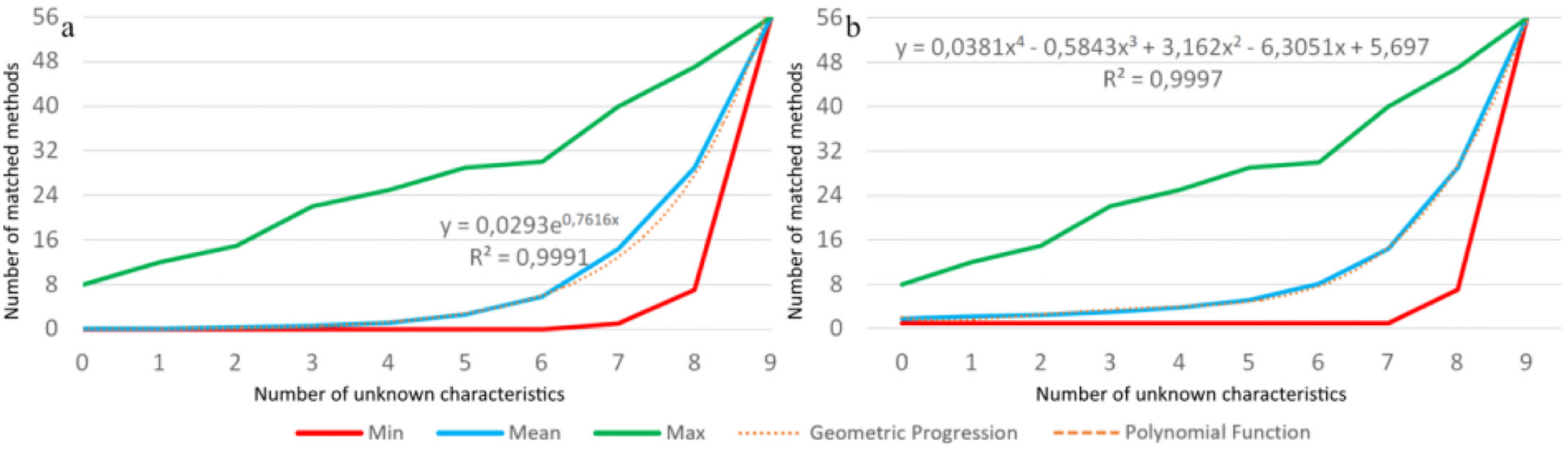

**Fig. 4.** Minimal, mean and maximal number of matching methods depending on the number of unknown characteristics for a 3-level hierarchy of classifiers, including (a) and excluding (b) the rules returning empty sets of methods.

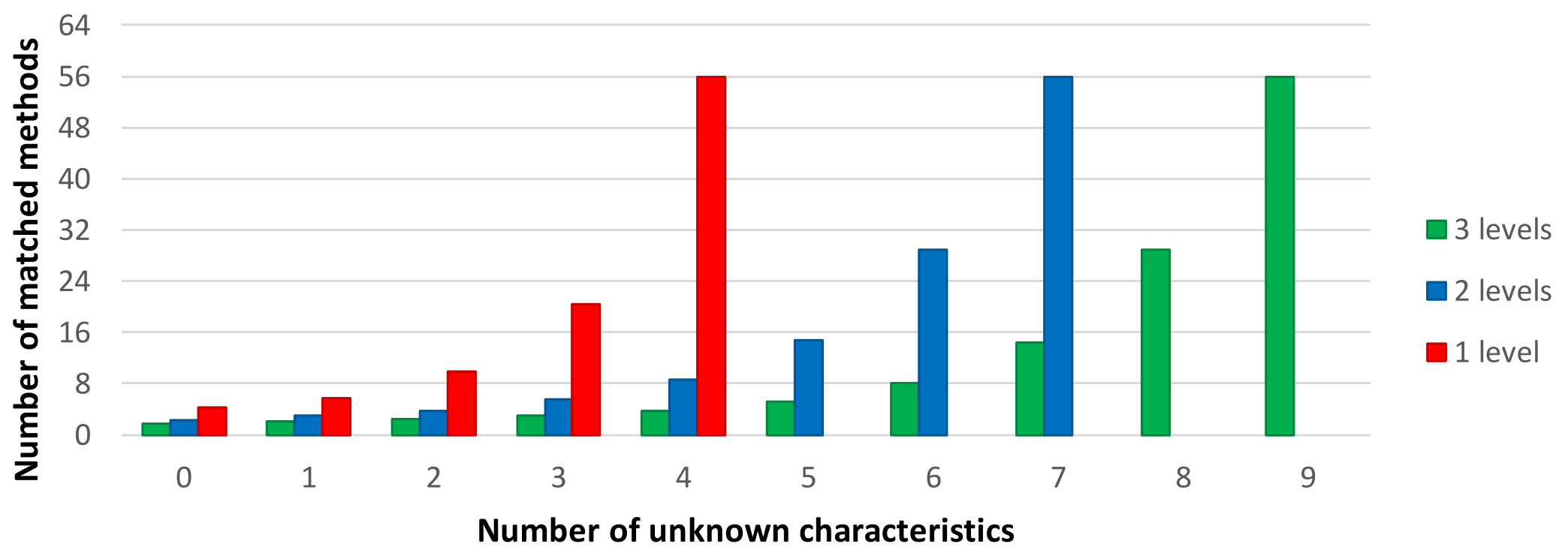

**Fig. 5.** Average number of methods returned by the rules on each level of classifiers' hierarchy depending on the number of unknown classifiers.

increase of the matching methods' count can be observed along with the increase of the number of unknown method properties. Similarly to the 3-level case, a significant increase of the number of methods can be observed at the expense of the number of rules.

In case of the single-level sequence of classifiers (Fig. 3a), the decision maker needs to take into account that even a single unknown value of the decision problem classifier results in a numerous set of rules and the count of rules ranging from 1 to as much as 30. It should be noted, however, that out of all 30 rules for a single-unknown scenario, only 7 of the rules stand out by providing 10 or more methods (19.5714 in average), whereas for the remaining rules 3 is the mode of the set. If the decision maker cannot produce two values of the methods' properties, the average number of methods returned grows to 9.8824. Moreover, if the DM can produce only a single value of the MCDA methods' classifiers, the average number of methods produced grows to 20.3636 with the minimum value of 1 and maximum value of 47 methods.

The aforementioned facts confirm that the introduction of additional levels to the rule set hierarchy largely increases its precision (Fig. 5), with correlation between the rule sets remaining at the very high level of 0.97 between 1-level and 2-level set or between 1-level and 3-level and 1 between the 2-level and 3-level set. This confirms the fact that the three rule sets can be used interchangeably, when a higher precision is expected.

Moreover, to verify the decision support abilities of the proposed framework, a methods' selection algorithm was developed (Supplementary material – Section 7). A prototype version of an online application supporting the MCDA method selection process handling uncertainty in decision problem description was implemented and is available at dedicated website at http://www.mcda.it.

## 4. Practical confirmation of the framework

The verification of the proposed framework was conducted with the use of decision making problems from the area of sustainable transport and logistics. The area of applications of multi criteria decision analysis methods is widely discussed in relation to sustainability assessment [30]. The focus of decision support systems dedicated for sustainable logistics is emphasized as well as the need for proper methods selection [34]. Various applications of the MCDA methods area observed in this field due to many conflicting criteria in the area of sustainable transport development [105], sustainable logistics practices [106], green supplier selection [107], green supply chain management [94], selection of transport technologies [108,109] or alternative fuels evaluation [110]. Cinelli et al. [30] emphasizes that various information types including uncertain parameters are required to perform sustainability assessment. To perform validation of proposed framework the set of reference cases of MCDA usage from above areas was prepared and is presented in Table 5. The examples include mainly problems of sustainable supplier selection, supply chain management, location choice, performance evaluation in green SCM, transport infrastructure design, alternative fuels selection, reverse logistics, sustainability assessment of urban systems with the focus on innovations. Each reference case is treated as expert recommendation to solve particular problem with specific MCDA method and is compared with the result delivered by the proposed framework. The obtained results are presented in Table 5.

Table 5 shows a high level of conformity between a selected MCDA method (literature source) and the results obtained by using the proposed framework. A few considered cases need additional clarification. This means a situation, where the proposed framework cannot assign any MCDA method (seven cases) or the characteristic of the considered problem causes that the wrong method is chosen (two cases). The framework does not return any methods for cases number: 5 [95], 12 [92], 16 [89], 29 [121], 32 [110], 33 [123] and 36 [125] (see Table 5):

- In cases 5 [95] and 16 [89], the AHP method was used only to determine relative importance of individual criteria and not to compare decision variants. A value greater than 0 is returned only by properties $m_{i1}$, and $m_{i1.1}$. All properties referring to decision variants and their comparisons obtain the value 0, including the most basic property determining whether decision variants are compared.
- As for the case 29 [121] the problem is that in this case, authors use equal weights to all the criteria of what was formally interpreted as a lack of weight. Properties $m_{i1}$ and $m_{i1.1}$ returned the value 0 instead of 1 and 2 respectively. Accordingly, the rule base fails to identify a suitable method, since it does not cover the situation where the weights are not used within the methods where it is possible to assign them.
- For cases 32 [110] and 33 [123], in which the AHP method was used to solve a problem, lack of choice of the method results from determining weights of criteria in an uncharacteristic manner. In the article [110], a sensitivity analysis was carried out and weights of criteria were expressed on a percentage scale. On the other hand, in case 33 [123], weights were attributed directly by the decision-maker and expressed on a point scale. In these cases, pairwise comparison matrices (along with a nine-degree Satty's scale) were not used in order to determine weights of criteria, the weights were not relative. Consequently, the property $m_{i1.1}$ returned the value 2 instead of 3.
- In Norese and Carbone [125] (case 36), criterial performances were qualitative not quantitative. That is why, property $m_{i2}$ assumed the value 1 instead of 2. As a result, for a given decision problem, a set of methods $\{E_T\}$ was not assigned.
- In example 12 [92] the wrong adjustment stems from the fact that the authors failed to use indifference and preferences

**Table 5**
Practical verification of decision rules with respect to the use of referential sources.

| No. | Particular MCDA problem descriptors | | | | | | | | | Description of the problem | The used MCDA method | The activated rule | Recommended subset of MCDA methods | Reference |
|---|---|---|---|---|---|---|---|---|---|---|---|---|---|---|
| | $c_1$ | $c_{1.1}$ | $c_2$ | $c_3$ | $c_{3.1}$ | $c_{3.1.1}$ | $c_{3.1.2}$ | $c_4$ | $c_{4.1}$ | | | | | |
| **1** | 1 | 2 | 2 | 1 | 2 | 0 | 1 | 1 | 0 | A model of selection the best innovation policies based on a number of criteria reflecting sustainability issues. | $\mathbf{E_S}$ | $R_{17}$ | $\{T_C\}$ | [101] |
| **2** | 1 | 3 | 2 | 1 | 1 | 1 | 0 | 3 | 2 | An integrated approach of fuzzy analytical hierarchy process (fuzzy AHP) and TOPSIS in evaluating the performance of global third party logistics service providers. | $\mathbf{A_F + T_P}$ | $R_{27}$ | $\{\mathbf{A_F + T_P}\}$ | [63] |
| **3** | 1 | 3 | 3 | 0 | 0 | 0 | 0 | 3 | 2 | Green supplier selection for an automobile manufacturing firm using AHP method. | $\mathbf{A_H}$ | $R_{30}$ | $\{\mathbf{A_H}, A_N, M_B, D_M, R_M\}$ | [107] |
| **4** | 1 | 3 | 3 | 0 | 0 | 0 | 0 | 3 | 2 | A performance evaluation model for the operations of the supply chain of an organization of the refrigeration equipment sector. | $\mathbf{M_B}$ | $R_{30}$ | $\{A_H, A_N, \mathbf{M_B}, D_M, R_M\}$ | [111] |
| **5** | 1 | 3 | 0 | 0 | 0 | 0 | 0 | 0 | 0 | The aim of the study is to investigate and to rank the pressures for GSCM based on experts' opinion using an Analytical Hierarchy Process (AHP) in the mining and mineral industry context. | $\mathbf{A_H}$ | – | ∅ | [95] |
| **6** | 1 | 3 | 3 | 0 | 0 | 0 | 0 | 3 | 2 | Choice of strategy for dealing with defective equipment in reverse logistics. | $\mathbf{A_N}$ | $R_{30}$ | $\{A_H, \mathbf{A_N}, M_B, D_M, R_M\}$ | [112] |
| **7** | 1 | 3 | 3 | 1 | 1 | 3 | 0 | 3 | 2 | Assessment of alternative suppliers for a business. | $\mathbf{A_{NF}}$ | $R_{31}$ | $\{A_F, \mathbf{A_{NF}}\}$ | [113] |
| **8** | 1 | 2 | 2 | 1 | 2 | 0 | 3 | 3 | 2 | Evaluation of energy business cases implemented in the North Sea Region and strategy recommendations using PROMETHEE II method. | $\mathbf{P_2}$ | $R_{21}$ | $\{\mathbf{P_2}\}$ | [114] |
| **9** | 1 | 3 | 3 | 0 | 0 | 0 | 0 | 3 | 2 | Choice of scenario for changes of used fuel for transportation. | $\mathbf{A_H}$ | $R_{30}$ | $\{\mathbf{A_H}, A_N, M_B, D_M, R_M\}$ | [115] |
| **10** | 1 | 3 | 3 | 0 | 0 | 0 | 0 | 3 | 2 | Choice of urban bypass project. | $\mathbf{A_H}$ | $R_{30}$ | $\{\mathbf{A_H}, A_N, M_B, D_M, R_M\}$ | [94] |
| **11** | 1 | 3 | 2 | 0 | 0 | 0 | 0 | 3 | 2 | Choice of fuel for public transport vehicles. | $\mathbf{A_H + T_P}$ | $R_{26}$ | $\{\mathbf{A_H + T_P}, A_H + V_K\}$ | [56] |
| **12** | 0 | 0 | 1 | 0 | 0 | 0 | 0 | 3 | 1 | Ranking of logistics platforms. | $\mathbf{E_4}$ | – | ∅ | [92] |
| **13** | 1 | 3 | 2 | 1 | 1 | 3 | 0 | 3 | 2 | Ranking of knowledge management solutions adopted in supply chain management. | $\mathbf{A_F + T_F}$ | $R_{29}$ | $\{\mathbf{A_F + T_F}, A_{NF} + T_F\}$ | [55] |
| **14** | 1 | 2 | 2 | 1 | 1 | 3 | 0 | 3 | 2 | The choice of location for urban distribution centers. | $\mathbf{T_F}$ | $R_{16}$ | $\{S_F, \mathbf{T_F}, V_F\}$ | [91] |
| **15** | 1 | 2 | 2 | 1 | 1 | 3 | 0 | 3 | 2 | A multi-criteria framework for comparative assessment of energy technologies in road transport taking into account technologies in terms of their environmental and economic impacts. | $\mathbf{T_F}$ | $R_{16}$ | $\{S_F, \mathbf{T_F}, V_F\}$ | [108] |
| **16** | 1 | 3 | 0 | 0 | 0 | 0 | 0 | 0 | 0 | The decision-making model handling relationships between GSCM practices and performances based on DEMATEL method. | $\mathbf{D_M}$ | – | ∅ | [89] |
| **17** | 1 | 3 | 2 | 0 | 0 | 0 | 0 | 3 | 2 | MCDA based system for the best municipal solid waste management scenario selection. | $\mathbf{A_H + V_K}$ | $R_{26}$ | $\{A_H + T_P, \mathbf{A_H + V_K}\}$ | [88] |
| **18** | 1 | 3 | 3 | 1 | 1 | 3 | 0 | 3 | 2 | Performance measurement of reverse logistics for the battery manufacturer. | $\mathbf{A_F}$ | $R_{31}$ | $\{\mathbf{A_F}, A_{NF}\}$ | [97] |
| **19** | 1 | 2 | 2 | 1 | 1 | 3 | 0 | 3 | 2 | Supplier choice of equipment from the customer to the manufacturer (reverse supply chain). | $\mathbf{T_F}$ | $R_{16}$ | $\{S_F, \mathbf{T_F}, V_F\}$ | [116] |
| **20** | 1 | 3 | 3 | 0 | 0 | 0 | 0 | 3 | 2 | Applying the analytic hierarchy process to the offshore outsourcing location decision. | $\mathbf{A_H}$ | $R_{30}$ | $\{\mathbf{A_H}, A_N, M_B, D_M, R_M\}$ | [117] |

(*continued on next page*)

**Table 5** (*continued*)

| No. | Particular MCDA problem descriptors | | | | | | | | | Description of the problem | The used MCDA method | The activated rule | Recommended subset of MCDA methods | Reference |
|---|---|---|---|---|---|---|---|---|---|---|---|---|---|---|
| | $c_1$ | $c_{1.1}$ | $c_2$ | $c_3$ | $c_{3.1}$ | $c_{3.1.1}$ | $c_{3.1.2}$ | $c_4$ | $c_{4.1}$ | | | | | |
| **21** | 1 | 2 | 2 | 1 | 1 | 3 | 0 | 3 | 2 | Assessment of balanced supplier performance (Green SCM). | $\mathbf{T_F}$ | $R_{16}$ | {$S_F$, $\mathbf{T_F}$, $V_F$} | [9] |
| **22** | 1 | 2 | 2 | 1 | 2 | 0 | 3 | 3 | 1 | Evaluation of alternative transport solutions for the urban transport system. | $\mathbf{E_3}$ | $R_{20}$ | {$\mathbf{E_3}$, $P_1$} | [93] |
| **23** | 0 | 0 | 2 | 0 | 0 | 0 | 0 | 1 | 0 | Supply chain optimization. | $\mathbf{G_P}$ | $R_4$ | {$\mathbf{G_P}$} | [69] |
| **24** | 1 | 2 | 1 | 0 | 0 | 0 | 0 | 1 | 0 | Evaluation of the performance of national transport systems in terms of impact on the economy, environment and society. | $\mathbf{E_1}$ | $R_{11}$ | {$\mathbf{E_1}$} | [96] |
| **25** | 1 | 3 | 3 | 0 | 0 | 0 | 0 | 3 | 2 | A decision to build a second airport in the metropolis. | $\mathbf{A_H}$ | $R_{30}$ | {$\mathbf{A_H}$, $A_N$, $M_B$, $D_M$, $R_M$} | [118] |
| **26** | 1 | 2 | 2 | 1 | 1 | 3 | 0 | 3 | 2 | Decision-making model in reverse logistics addressing green issues. | $\mathbf{V_F}$ | $R_{16}$ | {$S_F$, $T_F$, $\mathbf{V_F}$} | [119] |
| **27** | 1 | 2 | 2 | 1 | 1 | 3 | 0 | 3 | 2 | Sustainability assessment of urban transportation systems under uncertainty. | $\mathbf{T_F}$ | $R_{16}$ | {$S_F$, $\mathbf{T_F}$, $V_F$} | [98] |
| **28** | 1 | 3 | 2 | 1 | 1 | 3 | 0 | 3 | 2 | Green supplier evaluation. | $\mathbf{A_{NF}+T_F}$ | $R_{29}$ | {$A_F+T_F$, $\mathbf{A_{NF}+T_F}$} | [120] |
| **29** | 0 | 0 | 2 | 1 | 2 | 0 | 3 | 2 | 0 | A two-phase decision-making model integrating design and management of a Supply Chain from an outcome-driven perspective. | $\mathbf{E_T}$ | – | ∅ | [121] |
| **30** | 1 | 2 | 2 | 1 | 2 | 0 | 3 | 3 | 2 | Selection of the best sustainable concept using PROMETHEE II method. | $\mathbf{P_2}$ | $R_{21}$ | {$\mathbf{P_2}$} | [99] |
| **31** | 1 | 3 | 2 | 0 | 0 | 0 | 0 | 3 | 2 | A benchmarking framework evaluating the cold chain performance of a company. | $\mathbf{A_H+T_P}$ | $R_{26}$ | {$\mathbf{A_H+T_P}$, $A_H+V_K$} | [122] |
| **32** | 1 | 2 | 3 | 0 | 0 | 0 | 0 | 3 | 2 | An evaluation of alternative fuels for the road transport sector taking into account cost and policy criteria. | $\mathbf{A_H}$ | – | ∅ | [110] |
| **33** | 1 | 2 | 3 | 0 | 0 | 0 | 0 | 3 | 2 | MCDA based methodology for the Flanders in Action Process. | $\mathbf{A_H}$ | – | ∅ | [123] |
| **34** | 1 | 3 | 3 | 0 | 0 | 0 | 0 | 3 | 2 | An integrated balanced scorecard (BSC) and analytical hierarchy process (AHP) approach for supply chain management (SCM) performance evaluation. | $\mathbf{A_H}$ | $R_{30}$ | {$\mathbf{A_H}$, $A_N$, $M_B$, $D_M$, $R_M$} | [100] |
| **35** | 1 | 3 | 3 | 0 | 0 | 0 | 0 | 3 | 2 | An integrated model to evaluate Green Supply Chain's environmental performance. | $\mathbf{A_N}$ | $R_{30}$ | {$A_H$, $\mathbf{A_N}$, $M_B$, $D_M$, $R_M$} | [124] |
| **36** | 1 | 2 | 1 | 1 | 2 | 0 | 3 | 2 | 0 | An application of ELECTRE Tri to evaluate the airports' innovation. | $\mathbf{E_T}$ | – | ∅ | [125] |
| **37** | 1 | 2 | 2 | 0 | 0 | 0 | 0 | 3 | 2 | MCDA based tool supporting selection of the best biowaste management alternatives for stakeholders. | $\mathbf{T_P}$ | $R_{14}$ | {$E_V$, $M_U$, $M_V$, $S_A$, $S_M$, $\mathbf{T_P}$, $U_T$, $V_K$} | [126] |
| **38** | 1 | 3 | 2 | 0 | 0 | 0 | 0 | 3 | 2 | Sustainability-focused decision support system for supplier selection. | $\mathbf{A_H+V_K}$ | $R_{26}$ | {$A_H+T_P$, $\mathbf{A_H+V_K}$} | [127] |
| **39** | 1 | 3 | 2 | 1 | 1 | 2 | 0 | 3 | 2 | A fuzzy hierarchical TOPSIS based approach to evaluate different green initiatives and assess improvement areas when implementing green initiatives. | $\mathbf{A_H+T_F}$ | $R_{28}$ | {$\mathbf{A_H+T_F}$} | [128] |
| **40** | 1 | 2 | 2 | 1 | 2 | 0 | 3 | 3 | 2 | Logistics centre location choice. | $\mathbf{E_3}$ | $R_{21}$ | {$P_2$} | [129] |

thresholds, even though it is possible for the ELECTRE IV. Therefore, the properties $m_{i3}$, $m_{i3.1}$, and $m_{i3.1.2}$ were equal to 0.

Referring to the examples in which proposed framework selects a different method than the authors of the reference publication, we must consider the following examples: 1 [101] and 40 [129] (see Table 5):

- In the example 40 [129], the authors stated that they had received a total order of variants, while the ELECTRE III method allows only for a partial order. The property $m_{i4.1}$ mistakenly assumes a value of 2, so that, instead of the ELECTRE III method, PROMETHEE II is adapted (rule $R_{21}$ rather than $R_{20}$).
- The last situation, where used framework mistakenly matched the MCDA method applies to the example in row number 1 [101] from Table 5. The invalid assignment is caused by the fact that formally the authors used only the indifference threshold ($q$ and $p$ thresholds were equal). Therefore, the property $m_{i3.1.2}$ obtained value 1 instead of 3 and, consequently, rule $R_{17}$ was activated instead of rule $R_{18}$.

The study demonstrated usefulness of the proposed set of rules for selection of MCDA methods for real applications. The resulting accuracy of the selection is satisfactory. It should be noted that the occurrence of the missing or incorrect classifications was not a shortcoming of the proposed framework, but rather inadequacy of the problem analysis, often resulting from the inappropriate use of MCDA methods. Such applies even to popular methods such as AHP, where the publicly available algorithms were not always correctly implemented.

## 5. Summary

The selection of an MCDA method suitable for solving a specific decision problem is a vital element of the decision-making process. It is closely related to the issue of striving for the objectification of the decision support process itself [37]. A review of the literature confirms the dilemmas of researchers, and reveals that different methods are used for similar decision problems, leading to difficulties when comparing the results. The problem is addressed in various up-to-date studies, and the authors' conclusions clearly indicate the need for further studies in the search for generalized solutions independent of the current areas of usage of MCDA methods. Like it was showed, earlier approaches focused on MCDA selection problems only partially covered the problem because of limited set of considered methods and assumed precisely defined characteristics of the decision problem. Real situations are usually based on uncertain inputs not only at the level of detailed values of parameters but at more general specifications of decision problem as well.

The presented article contains a successful attempt to build a generalized MCDA method selection framework. The large-scale literature review allowed to extract a set of 56 up-to-date MCDA methods. Their profound analysis made it possible to identify sets of properties and to build on their basis a complete taxonomy of MCDA methods. It constituted the foundations for a formal presentation of the framework of the MCDA methods' selection in the form of a decision tree and a set of descriptors, as well as for the extraction of the set of decision rules. The presented framework for the selection of an MCDA method is based on the identified set of properties of the multi-criteria decision problem. The properties, which relate to decision problematics, comparison of variants, characterisation of weights, performance of alternatives, fuzzy data representation and aspects of imprecise in decision-makers' preferences, are used as a basis for the MCDA method selection.

The proposed framework constitutes formal guidelines for the selection of a particular MCDA method which is independent of the problem domain. While the earlier approaches for method selection take into account a limited number of methods, the current study uses the complete set of solutions available to date. It was shown in the research, that the hierarchical representation of the set of descriptors allowed the selection of MCDA methods also in situations of limited knowledge about the decision problem. The inclusion of uncertainty of the input data to the MCDA method selection rules in the presented approach allowed to address the issue of lack of knowledge in the description of the decision-making process. The modelling of the complete uncertainty space as a part of the proposed approach, enabled the analysis of the impact of the number of missing input data on the final form of the set of the recommended MCDA methods. These studies, along with the proposed framework, were also the basis for transferring the proposed solution to the public scope in the form of a complete, responsive, publicly available expert system supporting the selection of MCDA methods. The solution is available at http://www.mcda.it. When analyzing the effectiveness of the proposed approach, it is worth pointing out that in the empirical research the accuracy of recommendation of particular MCDA methods for a given decision-making situation was satisfactory.

Concluding, the main contributions of the work include:

- a generalised MCDA method selection framework for decision problems with theoretical background and wide applicability,
- the formal presentation of decision rules for MCDA method selection with the potential for direct application,
- guidelines for practitioners and a set of rules applicable in different areas of multicriteria decision making,
- hierarchical process of gathering knowledge about the decision problem from the decision maker,
- a complete analysis of the uncertainty influence on the final set of recommended MCDA methods,
- an algorithm for MCDA methods selection, as well as its implementation in a form of a web-based expert system.

In general, the presented framework provides a basis for construction of a knowledge database containing the rules for selection of a specific MCDA method from a set of all defined options, on the basis of detailed characteristics of the problem. The framework has some limitations, however. The presented set of rules in its current form is not always able to recommend a specific MCDA method. It can only propose a selection of potential methods. Additionally, due to the fact that the framework is based on the formal characteristics of a decision-making situation, the decision-making situation context aspect is omitted [20]. Consequently, the additional factors influencing the MCDA method selection for a given decision-making situation, such as the analyst's familiarity of the methods or the domain of the decision-making problem, were not studied.

The potential future works include the extension of the current collection of MCDA methods to include group decision making, and expansion of the database of reference cases. An additional challenging task would be the knowledge conceptualization and the construction of an ontology of decision problems and MCDA methods which would be used to select methods on the basis of classification results. This could lead to the development of a complete expert system supporting multi-criteria decision making. All the same, the authors would like to further improve their approach by adding support for more modern methods and adjusting the descriptors to match them. Therefore, the authors encourage and are looking forward to any forms of community input.

## Acknowledgment

This work was partially supported by the National Science Centre, Poland, grants no. 2017/25/B/HS4/02172 and 2017/27/B/HS4/01216.

## Supplementary materials

Supplementary material associated with this article can be found, in the online version, at doi:10.1016/j.omega.2018.07.004.

# Supplementary material

## **Section 1.** Characteristics of individual MCDA methods

| Method name | Essence of the method | Reference |
|---|---|---|
| AHP | The method allows the hierarchizing of a decision problem. Criterial evaluations of variants are set with reference to other variants in pairwise comparison matrices, with the use of nine-degree scale describing an advantage of one variant over another. Weights of criteria are set in a similar way. Next, the values of pairwise comparison matrices are aggregated to criterial preference vectors, then these are additively aggregated to a single synthesized criterion, on the basis of which a ranking of variants is determined. | [57] |
| ANP | The ANP method is a generalization of the AHP. Instead of hierarchizing a decision problem, the method allows constructing a net model, in which connections between criteria and variants, variant-criterion feedback or horizontal relations between individual criterion can occur. In this method, preference aggregation is based on a Markov chain and is carried out in a so-called Supermatrix. | [57] |
| ARGUS | Qualitative measurements are used in order to represent preferences on the ordinal scale. To compare variants with regard to criteria, five relations preferences are used, from indifference to very strong preference. Similarly, weights of criteria are expressed on a five-degree quality scale. Next, a preference graph is constructed using outranking relation, as it is the case in Electre I, for example. | [58] |
| COMET | Presentations are required for each criterion, triangle fuzzy numbers which determine degrees of variant affiliations to individual linguistic values describing individual criteria. Next, on the basis of values of vertexes of individual fuzzy numbers characteristic variants are generated. These variants are compared pairwise by a decision-maker and their model ranking is generated. | [59] |
| Electre I | It deals with the selection problematics. Preferences in Electre I methods are modelled by using binary outranking relations, and the method can be used when the criteria have been coded in numerical scales. The foundations of the method's algorithm are concordance and discordance indexes. Next, the best variants are those which are not outranked by any others. | [27] |
| Electre II | The method deals with the ranking problematics. The calculation algorithm is almost the same as in Electre I. However, in Electre II, one can distinguish between a strong and a weak preference relation. Computation procedures consist of four steps: partitioning set of variants, building a complete pre-order, determining a complete pre-order and defining the partial pre-order. | [27] |
| Electre III | It is based on pseudo-criteria (indifference, preference and veto thresholds are determined), instead of true criteria. After determining a decision-maker's preferences, concordance and discordance indexes are carried out, and the final ranking of variants is determined on the basis of results of distillation procedures. | [60] |
| Electre IS | Methodologically, it resembles Electre I and also deals with the selection problematics. The differences between the methods lie in the fact that in Electre IS pseudo-criteria (indifference, preference and veto thresholds) are used. | [27] |
| Electre IV | Similarly to Electre III, with regard to the use of pseudo-criteria. However, in Electre IV, one does not define weights of criteria, therefore, all criteria are equal. | [27] |
| Electre TRI | It deals with sorting problematics, but it uses pseudo-criteria. The method is very similar to Electre III in terms of procedures. In Electre Tri, decision variants are compared with variant profiles. The profiles are "artificial variants" limiting individual quality classes. The profiles are defined by a decision-maker, while determining the values of thresholds and weights of criteria. | [27] |
| EVAMIX | It allows both quantitative and qualitative criteria with the use of two separate domination measurements to be taken into consideration. Subsequent measurements are aggregated into one value describing the performance of a variant. Therefore, it is essential to apply a function which brings quantitative and qualitative measurements to the same level, and allows presenting them on a common scale. Weights of criteria are expressed in a quantitative way in this method. | [61] |
| Fuzzy AHP | A fuzzy version of the AHP method, in which criterial evaluations of variants are determined, with regard to other variants in pairwise comparison matrices, with the use of a fuzzy scale. | [62] |
| Fuzzy ANP | A fuzzy version of the ANP method, in which criterial evaluations of variants are determined with regards to other variants in pairwise comparison matrices, with the use of a fuzzy scale. | [64] |
| Methods of extracting the minimum and maximum values of the attribute | In the method, variants fulfilling individual criteria are chosen. However, all criteria have the same weight. In the method, for extracting the maximum value of the attribute one chooses variants maximally fulfilling one of the criteria. | [65] |
| Fuzzy PROMETHEE I | A fuzzy version of the Promethee I method. Weights and criterial variant evaluations are presented by means of fuzzy numbers. | [66] |
| Fuzzy PROMETHEE II | A fuzzy version of the Promethee II method. Weights and criterial variant evaluations are presented by means of fuzzy numbers. | [66] |
| Fuzzy SAW | A fuzzy version of the SAW method, in which weights and criterial variant evaluations are presented by means of triangle or trapezoid fuzzy numbers. | [67] |
| Fuzzy TOPSIS | The fuzzy TOPSIS method, similarly to its crisp version, is based on the concept of representing the positive ideal solution (PIS), negative ideal solution (NIS) and all variants on an Euclidean space. However, in this variant of the method, the values of decision attributes are represented by triangular or trapezoidal fuzzy numbers. The distance between the alternatives and PIS and NIS are calculated as | [53] |
| | a sum of distances between two fuzzy numbers representing each criterion separately. | |
| Fuzzy VIKOR | A fuzzy version of the VIKOR method, in which criterial variant evaluations are presented by means of triangle fuzzy numbers. | [68] |
| IDRA | IDRA (Intercriteria Decision Rule Approach) uses a mixed utility function which considers weights and compromises between criteria. On this basis, preference indexes depicting binary relations are determined. Aggregation of such indexes is based on the assumption that each piece of information about mutual substitutions between criteria constitutes a decision rule. A decision-maker assigns a different reliability to such a rule. Next, these reliabilities are used when calculating an aggregated preference index. | [13] |
| Lexicograph-ic method | It assumes both quantitative and qualitative assessments of variants, which are relative to subsequent criteria and ordinal criterion weights. In the next steps, variants are considered with relations to crite-ria ranking from the most to the least important one. During each step, a set of variants is reduced in such a way that there only variants are left, and these are, to a considerable degree, considered in a given step of a criterion. | [70] |
| MACBETH | Individual variants are here compared in a pairwise comparison matrix with the use of a seven-degree qualitative scale. Next, the comparison results are transformed into an interval scale. Weights of criteria, which are transferred from the qualitative scale to a quantitative one, and normalized to 100 (percentage scale), are determined in a similar way. Criterial preferences of variants are additively aggregated as a weighted average. | [71] |
| MAPPAC | It uses qualitative variant evaluations and quantitative, normalized to 1, weights of criteria. Variants' evaluations, which are relative to each criterion, are also normalized in the way where the best variant receives 1, the worst – 0, and other variants are given in-between values. For each pair of criteria, another matrix containing preference indexes is determined. All such matrices are aggregated into a global matrix, from which next preference rankings are constructed. | [72] |
| MAUT | In MAUT, the most important is an assumption that a decision-maker's preferences can be expressed by means of an analytical global utility function, while taking into consideration all considered crite-ria. Knowledge of this function makes it possible to obtain a set, ranked in terms of 'optimality,' of decision variants. | [73] |
| MAVT | MAVT is very similar to MAUT, but it uses the value function. In other words, in MAUT a ranking of variants is determined with the use of the utility function in the additive form, whereas in MAVT a ranking is obtained with the use of the multiplicative value function. | [73] |
| Maximax | It is based on the assumption that the productivity of a variant is as good as its strongest attribute. In this method, therefore, a variant characterized by the best value of a criterion (max) is chosen, with regard to which this variant is the strongest (max). | [74] |
| Maximin | It is based on the assumption that the productivity of a variant is as good as its weakest attribute. In this method, therefore, a variant characterized by the best value of a criterion (max) is chosen, with regard to which this variant is the weakest (min). | [74] |
| Maximin fuzzy method | A fuzzy version of the Maximin method, in which one operates on linguistic values. | [54] |
| MELCHIOR | In MELCHIOR pseudo-criteria (indifference and preference thresholds) are used. This method is similar to Electre IV in terms of calculation; however, in MELCHIOR an ordinal relation is set be-tween criteria. Preference aggregation takes place by testing conditions of concordance and lack of discordance. | [75] |
| Fuzzy meth-ods of ex-tracting the minimum and maximum values of the attribute | In the method, variants satisfactorily fulfilling individual criteria on the basis of fuzzy (linguistic) values are chosen. All criteria have the same weights. In order to extract the maximum value of the attribute, one may choose variants maximally fulfilling one of the criteria, on the basis of fuzzy (lin-guistic) values. | [74] |
| NAIADE I | As far as computation is concerned, NAIADE I is similar to Promethee, since variant ranking is de-termined on the basis of input and output preference flows. Nevertheless, when comparing variants there are six preference relations, which are based on trapezoid fuzzy numbers (apart from indiffer-ence of variants, one can distinguish weak and strong preference). In NAIADE I method, weights of criteria are not defined. | [76] |
| NAIADE II | Akin to NAIADE I, preference relations are defined on the basis of trapezoid fuzzy numbers. As in Promethee II, it allows a complete ranking of variants on the basis of net preference flows to be de-termined. | [76] |
| ORESTE | It requires presenting variant assessment and a ranking of criteria on an ordinal scale. Next, with the use of the distance function a complete order of variants, with regard to subsequent criteria is deter-mined, in which preference and indifference situations are acceptable. In the last step, rankings are aggregated into a global ranking, thus allowing an incomparability situation as well. | [77] |
| PACMAN | For every considered pair of criteria, the method allows distinguishing between a compensating (ac-tive) criterion and a compensated (passive) one. Separating active and passive compensation effect makes it possible to indicate compensation asymmetry. After analysing the compensation, a construc-tion of binary indexes, based on the evaluation of the degree of active and passive preferences, is performed. On this basis preference, indifference or incomparability relations between variants are determined. | [78] |
| PAMSSEM I | PAMSSEM I is a combination of Electre III, NAIADE I and Promethee I. Akin to NAIADE I, PAMSSEM I makes it possible to use fuzzy evaluations of variants. On the other hand, as in Electre III, there is a preference aggregation with the use of concordance and discordance indexes. Indiffer-ence, preference and veto thresholds are also used. Finally, as in Promethee I, a final ranking of vari-ants is obtained, with the use of input and output preference flows. | [79] |
| PAMSSEM II | A computational procedure is similar to PAMSSEM I; however, as in Promethee II, net preference flows are determined. | [79] |
| PRAGMA | It is based on MAPPAC and to a great extent; its course is identical to the MAPPAC algorithm. The only difference appears on the stage of aggregation of preferences from matrices of preference indexes determined for individual pairs of criteria. On this stage, for each individual matrix, a frequency matrix of partial rankings is determined. On their basis, a frequency matrix of a global ranking is calculated. Then, on the basis of this frequency matrix, preference aggregation into global rankings takes place. | [80] |
| PROMETHEE I | The Promethee method allows for determining a synthetic ranking of variants. Depending on implementation, the method can operate on real criteria or pseudo-criteria. Input and output preference flows are determined, on the basis of which one can create a partial Promethee I ranking can be created. | [81] |
| PROMETHEE II | In Promethee II, on the basis of input and output preference flows, the values of net preference flows for individual variants are calculated. On this basis, a complete ranking of variants is obtained. | [81] |
| QUALIFLEX | It allows use of qualitative evaluation variants, as well as quantitative and qualitative weights of criteria. Next permutations of rankings of variants are generated, and for every permutation; for every ranking, a concordance and discordance index is determined, firstly for individual ranking at the level of single criteria, then with relation to all possible rankings. As a result, an ordinal ranking with the best values of the concordance and discordance index is selected. | [82] |
| REGIME | It is based on the analysis of variants' concordance. A concordance matrix is determined, then three-valued pieces of information with positive/negative domination or variant indifference with regard to each criterion. Next, the probability of dominance for each compared pair of variants is determined, and on this basis, a variant order is obtained. | [83] |
| Simple Additive Weighting (SAW) | It requires giving quantitative evaluations of variants and weights. The evaluations of variants with regard to individual criteria should be proportionally normalized to the highest evaluation regarding each of the criteria. Preference aggregation comes down to determining a product of weights of a criterion, and the evaluation of a variant regarding this criterion. Next, all such products for a given variant are added up. | [74] |
| SMART | Criterial values of variants are calculated to a common internal scale. It is mathematically carried out by a decision-maker, and the value function used here (similarly to MAVT) using a range between the lowest and the highest values of variants for a given criterion. Variant evaluation here is not dependent on other ones. Moreover, the weights of criteria are determined by attributing explicit values to them, and not values relating to another criterion. | [84] |
| TACTIC | It is based on quantitative evaluations of variants and criteria weights. Furthermore, it allows applying real criteria and quasi-criteria and, in consequence, it uses the indifference and veto thresholds. As in Electre I and ARGUS, preference aggregation in this method is based on the analysis of concordance and discordance. | [15] |
| TOPSIS | TOPSIS (Technique for Order Performance by Similarity to Ideal Solution) is a method based on the concept of representing all the variants, along with the positive ideal solution (PIS) and the negative ideal solution (NIS) as points on an Euclidean space. The ranking of the variants is obtained based on the relative distances of the solutions from the reference PIS and NIS points. The best variant should have the smallest distance from the PIS and biggest distance from NIS. | [85] |
| UTA | A decision-maker's preferences are singled out from a reference set of variants. This set contains a list of sample decision options, which are ranked by the decision-maker in a ranking from the best to the worst one. In this ranking, a relation of preference or indifference can be used. Next, on the basis of the reference set fragmentary utility function for each criterion are created. Determining a global utility consists in adding values of all fragmentary utility functions for a given variant. | [86] |
| VIKOR | The VIKOR method consists in looking for a compromise, which is closest to an ideal solution. Firstly, the worst and the best values of individual criteria are determined. On the basis of the values, utility measures and regret measures for variants with regard to each criterion are calculated. Next, each variant's minimal and maximum values, whose aggregation allows determining the position of a variant in the ranking. | [87] |
| DEMATEL | The DEMATEL (Decision Making Trial and Evaluation Laboratory) method facilitates the decision making by providing a hierarchical structure of the criteria; however, contrary to the AHP method, it assumes that the elements of the structure are interdependent. A group of respondents is requested to evaluate the direct influence between any two factors on a 4-point scale, where 0 denotes no influence and 3 represents high influence. As a result of the method, a digraph showing casual relations among analyzed criteria is generated. | [89] |
| REMBRANT | The REMBRANDT (Ratio Estimation in Magnitudes or deci-Bells to Rate Alternatives which are Non-Dominated) technique is a multiplicative version of the AHP method. Pairwise comparisons between the objects are performed by the decision-maker on a geometric scale (1/16, 1/4, 1, 4, 16) where 1 denotes indifference, 4 and 16 represent weak and strict preference for the base object over the second object. The results of the comparisons are then converted into an integer-valued gradation index. As a result, the irrelative importance of the objects is determined. Finally, a subjective rank ordering of the objects is performed by aggregation of the results. | [90] |

**Supplementary material**
**Section 2.** Formal representation of decision problematics:

1) $\alpha$ decision problematics formal representation

$R_\alpha$ – solution of the $\alpha$ problematics

$$R_\alpha = f(w_\alpha)$$

where:

$$w_\alpha = \max_{w \in A' \subset A} \left\{ S_D(w): dim(w) = min \left\{ dim(v); \bigwedge_{v \in A \setminus A'} \bigvee_{q \in A'} \left( \neg S_D(q) \right) \right\} \right\}$$

dim – size,
$S_D()$ – DM’s satisfaction with the variant.

2) $\beta$ decision problematics formal representation

$R_\beta$ – solution of the $\beta$ problematics

$$R_\beta = u(w_B)$$

where:

$$\bigvee_{w \in A} w_B = w \wedge \bigwedge_{v \in A, v \neq w} \mu(w) > \mu(v)$$

$\mu$ – the norm related to the certain values.

3) $\gamma$ decision problematics formal representation

$R_\gamma$ – solution of the $\gamma$ problematics

$$R_\gamma = h(k_r)$$

where:

$$\bigvee_{k \in KR} K_r = k \wedge \bigwedge_{Kp \in KR, Kp \neq k} v(k) \approx v(kp)$$

$KR$ – the set of equivalence classes of variants from the set $A$,
$\approx$ – the relation of partial or complete order.

## Supplementary material

## **Section 3.** Approaches to selecting an MCDA method

| Type of approach | Aim | Highlights of the approach | Source |
|---|---|---|---|
| Benchmark | comparison | A solution of 4800 decision problems with the use of 8 MCDA methods. A comparison of the solutions with SAW results using 12 measurements and examining a rank reversal problem. | [19] |
| Benchmark | selection | Aggregation of 9 individual rankings into group rankings with the use of 18 fuzzy methods. Comparing solutions with all individual rankings using Spearman's correlation. | [25] |
| Benchmark | comparison | Generation of rankings with the use of 5 MCDA methods and their comparison with the use of Spearman's and Kendall's correlations. | [130] |
| Benchmark | comparison | A solution of three problems using 5 MCDA methods by experts. Comparison of individual methods by experts and comparison of the rankings obtained by the individual methods with the experts' rankings. | [131] |
| MCDA | selection | Selection of the best MCDA method (out of 13 methods) for a specific decision problem on the basis of the authors' criteria (out of 28). Aggregation of criterial preferences with the use of compromise programming. | [132,133] |
| MCDA | selection | Selection of the best MCDA method out of 15 for a specific decision problem by selected criteria (out of 49). Aggregation of criterial preferences with the use of compromise programming. | [134] |
| MCDA | selection | Selection of the best MCDA method out of 16 for a specific decision problem by Gershon's model as well as modified Tecle's model in which when aggregating preferences, a Promethee method was used. | [48] |
| MCDA | selection | Selecting a group of the best methods out of 24 MCDA ones as well as a management method for a specific decision problem. Ten criteria were used to construct a ranking of methods, and criterial preferences were aggregated with the use of fuzzy sets and a FAD (fuzzy axiomatic design) procedure. | [135] |
| MCDA | selection | Selection of the best out of 9 MCDA methods for eight individual decision problems. Selection of individual methods on the basis of 11 criteria. Preference aggregation with the use of fuzzy sets and an FAD-MSI (fuzzy axiomatic design – model selection interface) procedure. | [42] |
| MCDA | selection | Selection of the best out of 5 MCDA methods for eight individual problems defined by various contexts of decision situations. Preference aggregation with the use of fuzzy sets and an FAD-MSI procedure, on the basis of 10 criteria. | [136] |
| informal | selection | Selection of optimal methods out of 41 MCDA ones to solve a specific decision problem on the basis of 6 criteria formulated ad-hoc describing characteristics which an appropriate method should have. | [137] |
| informal | selection | Eleven characteristics of MCDA methods and their requirements were defined. Selection of methods, which have appropriate features to solve a given decision problem, out of 15 MCDA methods. | [138] |
| semiformal | comparison | Comparison of seven MCDA methods with regard to 21 characteristics. A distinction of the most important characteristics which are essential when applying a method in any given decision problem. Formulation of six guidelines concerning decision situations in which specific MCDA methods should be used. | [139] |
| semiformal | comparison | Comparison of five MCDA methods with regard to 10 characteristics defining, among other things, the capability of a method for sustainability assessment. | [30] |
| semiformal | comparison/ selection | Comparison of six MCDA methods and their groups with reference to 15 characteristics. Selection of an MCDA method for a given decision problem in a case study. | [50] |
| formal | comparison | Comparison of 17 MCDA methods on the basis of taxonomy taking into consideration the type of preference information and information features. Formulation of a decision tree for selecting an MCDA method on the basis of 7 characteristics / tree nodes. | [74] |
| formal | comparison /selection | Introduction of a taxonomy of methods with regard to characteristics of criterial assessment and weights of criteria. Comparison of 26 MCDA methods on the basis of 5 characteristics presented as nodes of a decision tree. Selection of an optimal method which could be used in a specific decision problem from among methods presented in a given leaf of a tree. | [140] |
| formal | comparison | Comparison of 29 MCDA methods with regard to 14 characteristics. Presentation of an outline of a concept of constructing an MCDA selection tree on the basis of presented characteristics. | [20] |
| formal | comparison | Comparison of 30 MCDA methods with regard to 10 characteristics as well as ten possible types of data describing criterial performance of input instances. As a result, 24 possible inputs to a decision process and seven outputs were formulated. Rules for selecting an MCDA method were introduced on the basis of a given input and an expected output of a decision process. | [51] |
| Formal | selection | Selection of the best MCDA methods out of 24 for 8 different decision problems. The selection made on the basis of 20 characteristics of a decision problem (an IDEA model), given by the decision-maker and their corresponding characteristics of individual methods. | [32] |
| Formal | selection | Selection of the best MCDA methods out of 21 for 2 different decision problems. The selection was made on the basis of 20 characteristics of a decision problem in accordance with an IDEA model (Integrated Decision Aid). | [22] |
| Formal | selection | Selection of an optimal MCDA method (out of 22) on the basis of a classifier (artificial neural networks) using 20 characteristics of an $IDEA_{ANN}$ model. Tests of the classifier with the division into learning and testing sets. | [21] |

## Supplementary material

**Section 4.** Analysis of the influence of the lack of knowledge about a particular level of classifiers on the quality of recommendation

A detailed analysis is presented in the following stages. Initially, the count of rules returned for each level of hierarchy was studied. In case of a single level, where a decision problem is described by four classifiers, a total of 48 rules is possible, returning 1.6667 methods on average. However, as it can be observed on Figure S4a, most of the rules provide empty sets of methods. If these rules are omitted, only 13 rules, i.e. 27% remain under consideration, with an average of 4.3077 methods returned. Nonetheless, it should be noticed from the analysis of Fig. 2a that only 3 rules return more methods than average. Therefore, it can be concluded that if only four classifiers are used, the decision maker (DM) would statistically obtain 2-5 methods, however, there is a possibility that a set of 14 or 16 methods can be matched.

On the other hand, if the complexity of the classifiers' hierarchy is increased to two levels (Figure S4b), a set of 240 rules can be obtained, returning 0.2333 methods on average. However, as much as 90% of the rules return empty sets of rules, and the remaining 25 rules return averagely 2.24 methods, i.e. approximately 5% less than in case of the single-level hierarchy. Only 5 rules return more than average number of rules, however, the most general rule returns only 8 methods, which is twice more precise than the most general rule for the four classifiers' case.

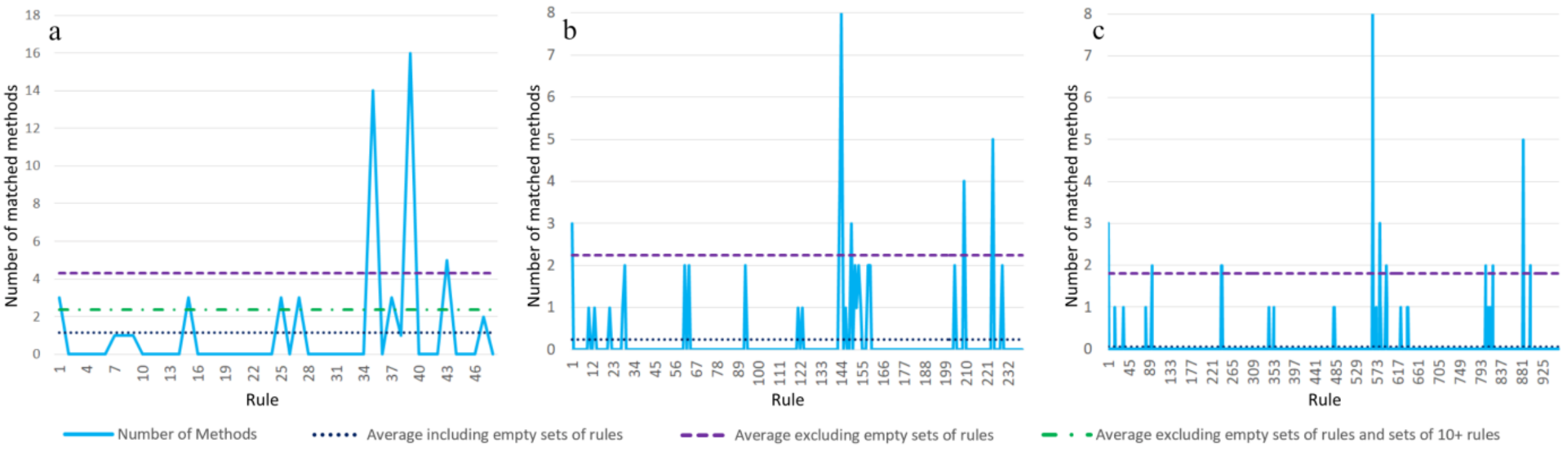

**Fig. S4.** Number of methods in the sets returned by each rule, depending on the structure of decision problem: a) single-level; b) two levels; c) three levels of criteria.

Finally, in case when 9 classifiers organized into 3 levels are considered (Figure S4c), a total of 960 rules can be produced, with only 31, i.e. 3% not being empty. The average number of methods returned by these non-empty rules is equal to 1.8065. Therefore, this approach provides the most precise rules.

**Supplementary material**
**Section 5.** Rules' extraction procedure.

**Table S5** The valid sets of values for rules extraction

| Step 1 | | Step 2 | Step 3 | | | | Step 4 | |
|---|---|---|---|---|---|---|---|---|
| $c_1$ | $c_{1.1}$ | $c_2$ | $c_3$ | $c_{3.1}$ | $c_{3.1.1}$ | $c_{3.1.2}$ | $c_4$ | $c_{4.1}$ |
| 0 | 0 | 1 | 0 | 0 | 0 | 0 | 1 | 0 |
| 1 | 1 | 2 | 1 | 1 | 1 | 0 | 2 | 0 |
| 1 | 2 | 3 | 1 | 1 | 2 | 0 | 4 | 0 |
| 1 | 3 | ? | 1 | 1 | 3 | 0 | 3 | 1 |
| 1 | ? | | 1 | 2 | 0 | 1 | 3 | 2 |
| ? | ? | | 1 | 2 | 0 | 2 | 3 | ? |
| | | | 1 | 2 | 0 | 3 | ? | ? |
| | | | 1 | 3 | 1 | 1 | | |
| | | | 1 | 3 | 1 | 2 | | |
| | | | 1 | 3 | 1 | 3 | | |
| | | | 1 | 3 | 2 | 1 | | |
| | | | 1 | 3 | 2 | 2 | | |
| | | | 1 | 3 | 2 | 3 | | |
| | | | 1 | 3 | 3 | 1 | | |
| | | | 1 | 3 | 3 | 2 | | |
| | | | 1 | 3 | 3 | 3 | | |
| | | | 1 | 1 | ? | 0 | | |
| | | | 1 | 2 | 0 | ? | | |
| | | | 1 | 3 | ? | ? | | |
| | | | 1 | 3 | ? | 1 | | |
| | | | 1 | 3 | ? | 2 | | |
| | | | 1 | 3 | ? | 3 | | |
| | | | 1 | 3 | 1 | ? | | |
| | | | 1 | 3 | 2 | ? | | |
| | | | 1 | 3 | 3 | ? | | |
| | | | 1 | ? | ? | ? | | |
| | | | ? | ? | ? | ? | | |

(? means unknown value)

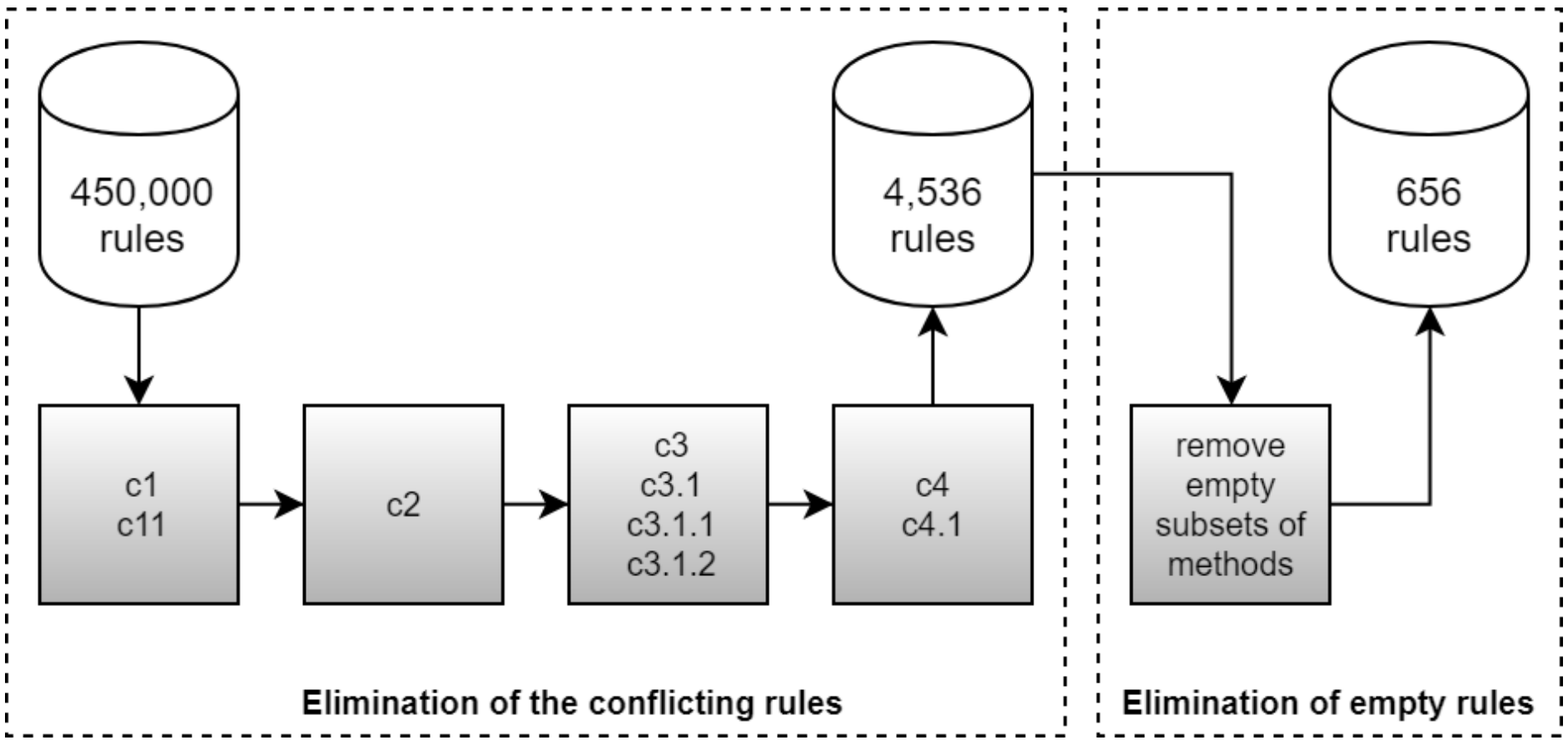

**Fig. S5** The process of extraction of valid rules from the full set of 450,000 rules.

**Supplementary material**

**Section 6.** Analysis of the rule sets when rules returning empty sets of methods are considered

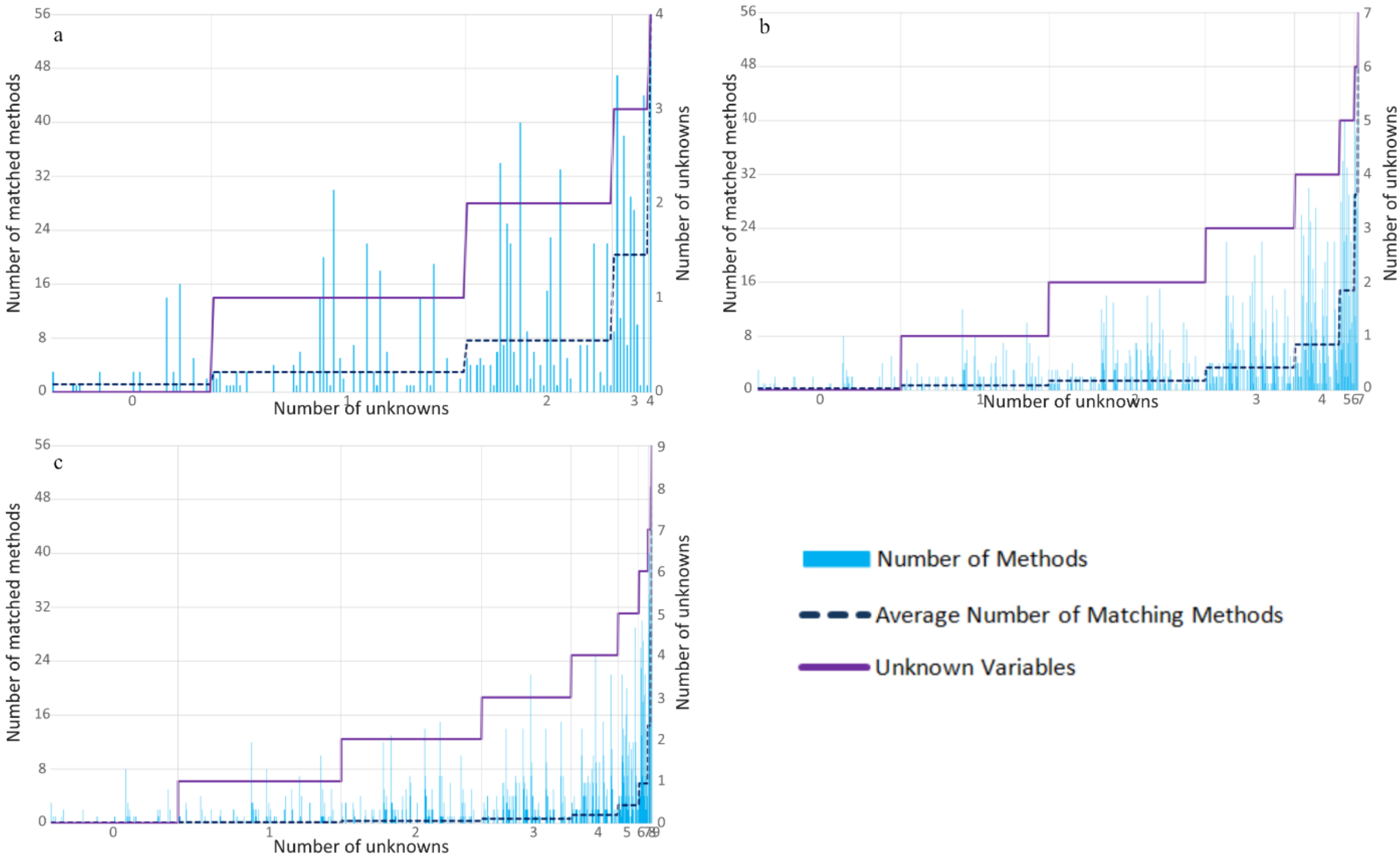

**Fig. S6.** A histogram-like analysis of the possible MCDA method selection rules depending on the number of unknown characteristics, in cases of a single level (a), two levels (b) and three levels (c) of MCDA methods' properties, including the rules returning 0 methods.

Figure S6 illustrates a histogram-like analysis of the possible MCDA selection rules, based on the number of unknown MCDA characteristics, as well as the complexity of the structure. The x-axis and right y-axis represent the number of the characteristics that are unknown, whereas the left y-axis represents the growth in number of methods that possibly meet these unknown characteristics. The bars in the chart represent the precise count of methods matching every single rule, whilst the dashed line represents their average count for the corresponding number of unknowns. All 4,536 rules are presented, including the ones returning empty sets of methods.

When Figure S6c is analyzed in detail, it can be observed that when a single property is unknown to the decision maker, the 3-level decision framework still allows to limit the matching number of MCDA methods to a range of 0 to 12, with average value equal to 0.1445 when all 1232 rules are considered.

If the number of unknown decision problem characteristics grows to two values, a slight decrease of their number is observed, to a total of 1060 rules. The average number of matched methods increases only slightly to 0.3113. A similar growth of possibly matching methods can be observed also when the count of unknown characteristics grows to 3 and 4, with the average equal to 0.6193 and 1.2147 respectively.

The average count of matching methods for cases when 5, 6, 7 or 8 of the total of 9 properties are unknown is equal to 2.6709, 5.9118, 14.4545 and 29 respectively. This growth can be mapped by a geometric progression with $R^2$ equal to 0.9991. It is also important to note from Figure S6c, that along with the increase of the count of methods, the number of rules decreases when more MCDA methods' properties are unknown.

Figures S6a, S6b depict the two remaining scenarios when the knowledge about the decision problem is structured only into two levels (Figure S6b) or into a sequence of 4 main classifiers (Figure S6a). In the former, if the number of unknown method properties remains in the range of 1 – 3, the increase of matching methods' count remains linear, however for 4 or more unknown classifiers it becomes exponential, with the average count of matching methods equal to 6.7467, 14.8 and 29 for 4, 5 and 6 unknown classifiers respectively. Similarly to the 3-level case, the significant increase of the number of methods can be observed at the expense of the number of rules.

In case of the single-level sequence of characteristics (Figure S6a), the decision maker needs to take into account that even a single unknown value of the decision problem classifier results in a numerous set of rules and the count of rules ranging from 0 to as much as 30. It should be noted, however, that out of all 76 rules for a single-unknown scenario, only 7 of the rules stand out by providing 10 or more methods (19.5714 in average). If the decision maker cannot produce two values of the methods' properties, the average number of methods returned grows to 7.6364. Moreover, if the DM can produce only a single value of the MCDA methods' properties, the average number of methods produced grows to 20.3636 with the minimum value of 1 and maximum value of 47 methods, thus confirming the exponential progression of the number of methods matched.

**Supplementary material**
**Section 7.** MCDA methods’ selection algorithm based on an uncertain set of classifiers’ values.

**Input:**

- values of classifiers c1 [0,1,?], c1.1 [0,1,2,3,?], c2 [1,2,3], c3[0,1,?], c3.1 [0,1,2,3,?], c3.1.1 [0,1,2,3,?], c3.1.2 [0,1,2,3,?], c4 [1,2,3,4,?], c4.1 [0,1,2,?];
- database of methods with assigned values of classifiers

**Output:**

- array of matching methods with the values of the classifiers for each of the method

```
$qb = new QueryBuilder();
if input[c1] != ‘?’ {
  $qb->addCondition(c1 = input[c1]);
}
if input[c1.1] != ‘?’ {
  if input[c1] == ‘?’ {
    throw new InvalidRequestException();
  }
  $qb->addCondition(c1 = input[c1]); // automatically set c1
  $qb->addCondition(c1.1 = input[c1.1]);
}
if input[c2] != ‘?’ {
  $qb->addCondition(c2 = input[c2]);
}
if input[c3] != ‘?’ {
  $qb->addCondition(c3 = input[c3]);
}
if input[c3.1] != ‘?’ {
  if input[c3] == ‘?’ {
    throw new InvalidRequestException();
  }
  $qb->addCondition(c3 = input[c3]); // automatically set c3
  $qb->addCondition(c3.1 = input[c3.1]);
}
if input[c3.1.1] != ‘?’ {
  if input[c3] != 1 && input[c3.1] not in [1,3] {
    throw new InvalidRequestException();
  }
  $qb->addCondition(c3 = input[c3]); // automatically set c3
  $qb->addCondition(c3.1.1 = input[c3.1.1]);
}
if input[c3.1.2] != ‘?’ {
  if input[c3] != 1 && input[c3.1] not in [1,3] {
    throw new InvalidRequestException();
  }
  $qb->addCondition(c3 = input[c3]); // automatically set c3
  $qb->addCondition(c3.1.2 = input[c3.1.2]);
}
```

```
if input[c4] != ‘?’ {
  $qb->addCondition(c4 = input[c4]);
}
if input[c4.1] != ‘?’ {
  if input[c4] == ‘?’ {
    throw new InvalidRequestException();
  }
  $qb->addCondition(c4 = input[c4]); // automatically set c4
  $qb->addCondition(c4.1 = input[c4.1]);
}
$methods = $qb->queryDatabase(); // use the collected data to query the database for results
return $methods;
```

**Supplementary material**

**Section 8.** Summary of MCDA method selection approaches

| References | Number of considered MCDA methods | Number of citations according WebofKnowledge |
|---|---|---|
| Moffett A, Sarkar S. Incorporating multiple criteria into the design of conservation area networks: a minireview with recommendations. Divers and Distrib 2006;12:125–37. | 26 | 82 |
| Al-Shemmeri T, Al-Kloub B, Pearman A. Model choice in multicriteria decision aid. Eur J of Oper Res 1997;97:550–60. | 16 | 40 |
| Moghaddam NB, Nasiri M, Mousavi SM. An appropriate multiple criteria decision making method for solving electricity planning problems, addressing sustainability issue. Int J of Environ Sci & Technol 2011;8:605–20. | 15 | 4 |
| Celik M, Cicek K, Cebi S. Establishing an international MBA program for shipping executives: Managing OR/MS foundation towards a unique curriculum design, IEEE; 2009, p. 459–63. | 24 | 1 |
| Celik M, Deha Er I. Fuzzy axiomatic design extension for managing model selection paradigm in decision science. Expert Syst with Appl 2009;36:6477–84. | 9 | 14 |
| Celik M, Topcu YI. Analytical modelling of shipping business processes based on MCDM methods. Marit Policy & Manag 2009;36:469–79. | 24 | 5 |
| Cicek K, Celik M. Multiple attribute decision-making solution to material selection problem based on modified fuzzy axiomatic design-model selection interface algorithm. Mater & Des 2010;31:2129–33. | 5 | 16 |
| Cicek K, Celik M, Ilker Topcu Y. An integrated decision aid extension to material selection problem. Materials & Design 2010;31:4398–402. | 21 | 8 |
| Cinelli M, Coles SR, Kirwan K. Analysis of the potentials of multi criteria decision analysis methods to conduct sustainability assessment. Ecol Indic 2014;46:138–48. | 5 | 48 |
| Guitouni A, Martel J-M. Tentative guidelines to help choosing an appropriate MCDA method. Eur J of Oper Res 1998;109:501–21. | 29 | 223 |
| Hajkowicz S, Higgins A. A comparison of multiple criteria analysis techniques for water resource management. Eur J of Oper Res 2008;184:255–65. | 5 | 109 |
| Ulengin F, Ilker Topcu Y, Onsel Sahin S. An Artificial Neural Network Approach to Multicriteria Model Selection. In: Köksalan M, Zionts S, editors. Multiple Criteria Decision Making in the New Millennium, vol. 507, Berlin, Heidelberg: Springer Berlin Heidelberg; 2001, p. 101–10. | 22 | 1 |
| Zanakis SH, Solomon A, Wishart N, Dublish S. Multi-attribute decision making: A simulation comparison of select methods. Eur J of Oper Res 1998;107:507–29. | 8 | 266 |

## Supplementary material

**Section 9.** Detailed analysis of the most cited MCDA method selection approaches

| MCDA method | Ref1 | Ref2 | Ref3 | Ref4 | Ref5 | Ref6 | Ref7 |
|---|---|---|---|---|---|---|---|
| AHP | V | V | V | | V | V | V |
| ANP | V | | | | | | |
| ARGUS | V | | V | | | | |
| COMET | | | | | | | |
| ELECTRE II | V | V | V | V | | V | V |
| ELECTRE II | V | V | V | | | V | V |
| ELECTRE III | V | V | V | | | V | V |
| ELECTRE IS | V | V | | | | V | V |
| ELECTRE IV | V | V | V | | | V | V |
| ELECTRE TRI | V | V | | | | V | V |
| EVAMIX | V | V | | V | | | |
| Fuzzy AHP | | | | | | | |
| Fuzzy ANP | | | | | | | |
| Fuzzy methods of extracting the minimum and maximum values of the attribute | | V | | | | | |
| Fuzzy PROMETHEE I | | | | | | | |
| Fuzzy PROMETHEE II | | | | | | | |
| Fuzzy SAW | | V | | | | | |
| Fuzzy TOPSIS | | | | | | | |
| Fuzzy VIKOR | | | | | | | |
| IDRA | V | | V | | | | |
| Lexicographic method | | V | | | | | |
| MACBETH | V | | V | | | | |
| MAPPAC | V | | V | | | | |
| MAUT | V | V | | | | V | V |
| MAVT | | V | V | | | | |
| Maximax | | | V | | | | |
| Maximin | | V | V | | | | |
| Maximin fuzzy method | | V | | | | | |
| MELCHIOR | V | V | V | | | | |
| Methods of extracting the minimum and maximum values of the attribute | | V | | | | | |
| NAIADE I | | V | | | | | |
| NAIADE II | | V | | | | | |
| ORESTE | V | V | V | | | | |
| PACMAN | V | | V | | | | |
| PAMSSEM I | | | | | | | |
| PAMSSEM II | | | | | | | |
| PRAGMA | V | | V | | | | |
| PROMETHEE I | V | V | V | | | V | V |
| PROMETHEE II | V | V | V | V | | V | V |
| QUALIFLEX | V | V | V | | | | |
| REGIME | V | V | V | | | | |
| Simple Additive Weighting (SAW) | | V | | V | | | |
| SMART | | V | | | | | |
| TACTIC | V | | V | | | | |
| TOPSIS | | V | V | | V | | |
| UTA | V | V | V | | | | |
| VIKOR | | | | | | | |
| DEMATEL | | | | | | | |
| REMBRANDT | | | | | | | |